# SFE: A Simple, Fast and Efficient Feature Selection Algorithm for High-Dimensional Data

Behrouz Ahadzadeh, Moloud Abdar, Fatemeh Safara, Abbas Khosravi, *Senior Member, IEEE*, Mohammad Bagher Menhaj, Ponnuthurai Nagaratnam Suganthan, *Fellow, IEEE*

*Abstract*— In this paper, a new feature selection algorithm, called SFE (Simple, Fast, and Efficient), is proposed for high-dimensional datasets. The SFE algorithm performs its search process using a search agent and two operators: non-selection and selection. It comprises two phases: exploration and exploitation. In the exploration phase, the non-selection operator performs a global search in the entire problem search space for the irrelevant, redundant, trivial, and noisy features, and changes the status of the features from selected mode to non-selected mode. In the exploitation phase, the selection operator searches the problem search space for the features with a high impact on the classification results, and changes the status of the features from non-selected mode to selected mode. The proposed SFE is successful in feature selection from high-dimensional datasets. However, after reducing the dimensionality of a dataset, its performance cannot be increased significantly. In these situations, an evolutionary computational method could be used to find a more efficient subset of features in the new and reduced search space. To overcome this issue, this paper proposes a hybrid algorithm, SFE-PSO (particle swarm optimization) to find an optimal feature subset. The efficiency and effectiveness of the SFE and the SFE-PSO for feature selection are compared on 40 high-dimensional datasets. Their performances were compared with six recently proposed feature selection algorithms. The results obtained indicate that the two proposed algorithms significantly outperform the other algorithms, and can be used as efficient and effective algorithms in selecting features from high-dimensional datasets.

## I. INTRODUCTION

THE development of computer systems and new technologies has led to the production and storage of large amounts of high-dimensional datasets in various applications such as healthcare, e-commerce, bioinformatics, social media, and transportation [1], [2], [3]. Machine learning algorithms are widely used for classification problems; however, the curse of dimensionality prevents achieving the desired result; because the data distributed in high-dimensional datasets adversely affects algorithms designed for low-dimensional datasets. In addition, the high dimensionality of data increases the time and space complexity of machine learning algorithms in data classification. To solve these problems that have a negative impact on the efficiency of machine learning algorithms, the dimensionality reduction methods are commonly proposed [3].

In general, dimensionality reduction methods are usually performed in the preprocessing phase of the entire classification. It can be divided into feature extraction and feature selection (FS). The main objective of feature extraction algorithms is to create new features based on the main features of the dataset and to reduce the dimensionality of the dataset. Newly constructed features are usually a linear or non-linear combination of the main features [4]. FS algorithms select a relevant subset of features that contain important information about the data and would result in better performance of machine learning algorithms. In addition, the unimportant features (i.e., irrelevant, redundant, trivial and noisy features) would be removed from the feature set [3], [5] because these features would cause overfitting and reduce the performance of the algorithms. FS algorithms are typically employed in the preprocessing phase to reduce the computational and space complexity of the algorithms, which increases the overall performance of machine learning algorithms. In addition, the generality of the algorithms would be improved.

The basis of feature extraction methods is the construction of new features based on the main features, consequently, the new features lack the physical meaning of the main features. Therefore, these methods are not used in applications such as text mining and genetic analysis [3]. In contrast, FS methods retain the physical meaning of the features as they directly select the relevant features that provide the correct interpretation for the stated applications. Furthermore, due to the high computational complexity of feature extraction methods in high-dimensional datasets, these methods may not achieve the expected performance in some applications. For the above-mentioned reasons, FS methods have recently been researched and improved [3], [4].

In general, FS algorithms can be divided into three categories: filter-based, wrapper-based, and embedded [3]. The typical filter-based feature selection algorithms consist of two main steps. In the first step, the importance and the rank of each feature are determined according to various criteria such as distance, correlation, consistency and information-gain. In the second step, features with a lower rank are then filtered [3], [6], [7]. However, some filter-based FS algorithms, such as the Correlation FS algorithm (CFS) [8], do not necessarily follow these two steps.

The typical wrapper-based FS algorithms use a machine learning classifier to evaluate the selected features. These methods generally involve two major steps. In the first step, an optimal subset of features is searched using search methods. In the second step, various classification methods such as K-Nearest Neighbor (KNN) and Support Vector Machines (SVM) are used for evaluating the optimal subset of the features [9],

*Corresponding author: Moloud Abdar (E-mail: m.abdar1987@gmail.com)*
Behrouz Ahadzadeh is with the Department of Electrical, Computer and IT Engineering, Qazvin Branch, Islamic Azad University, Qazvin, Iran (E-mail: b.ahadzade@yahoo.com)
Moloud Abdar and Abbas Khosravi, are with the Institute for Intelligent Systems Research and Innovation (IISRI), Deakin University, Australia (E-mail: m.abdar1987@gmail.com and abbas.khosravi@deakin.edu.au)
Fatemeh Safara is with the Department of Computer Engineering, Islamshahr Branch, Islamic Azad University, Islamshahr, Iran (E-mail: fsafara@yahoo.com)
Mohammad Bagher Menhaj is with the Department of Electrical Engineering Amir Kabir University of Technology Tehran, Iran (E-mail: menhaj@aut.ac.ir)
P. N. Suganthan is with the KINDI Center for Computing Research, College of Engineering, Qatar University, Doha, Qatar (E-mail: p.n.suganthan@qu.edu.qa).

[10]. Both steps are repeated to achieve the expected performance [11], [12]. However, some wrapper-based FS algorithms also do not necessarily follow these two steps. The embedded FS methods use classification algorithms for feature selection, similar to the wrapper-based methods, except that FS is embedded in the classifier structure [13]. One of the FS methods is the decision tree method [14]. The selected features are the features that exist in the final tree after training the decision tree [13], [15].

Some of the previous studies [3], [16], [17] divide the FS into four categories by including the hybrid FS methods. In hybrid FS methods, a combination of filter-based, wrapper-based, and embedded FS algorithms are utilized. In general, wrapper-based and embedded methods are more efficient than filter-based methods because they interact with classifiers in the FS process. In contrast, the computational cost of filter-based methods is less than that of wrapper-based and embedded methods. The embedded methods are usually less costly than wrapper-based methods. Though, these methods may only be used mostly with certain classification methods such as DT, SVM [3], [11], [18]. The hybrid methods also have acceptable performance due to taking advantage of different algorithms but these methods are still face with the challenge of high computational costs. Therefore, in order to achieve better efficiency in the FS process, we should reduce the computational complexity of wrapper-based methods which are more efficient method than filter-based and embedded methods. For this reason, the focus of this paper is to provide a wrapper-based FS method for high-dimensional data.

The wrapper-based FS methods use search algorithms to find the optimal subsets of features. The simplest method to search for an optimal subset of features in a dataset is to search the entire problem search space and evaluate the performance of each possible subset, which is called an exhaustive search method. In a dataset with $n$ features, the number of possible subsets is $2^n - 1$. Therefore, for large $n$, using a complete search to find the best subset of features is practically impossible even with today's most powerful computers [9], [18]-[20]. For this reason, instead of using an exhaustive search algorithm, researchers employed heuristic algorithms such as sequential forward selection (SFS) [21], sequential backward selection (SBS) [22], plus-minus-r selection method (LRS) [23], sequential floating forward selection (SFFS), and sequential floating backward selection (SFBS) algorithms [24]. Although, these algorithms can significantly reduce the time complexity of the complete search, they may trap at local optimal points because of their consistent strategy in the search process [9], [19]. In this regard, in recent years, researchers have used another class of search algorithms called evolutionary computational (EC) methods for feature selection. Simple implementation, global search capability, high flexibility to be used in various applications, near-optimal global solution in acceptable space and time complexity, and ability to escape local optimal points due to their stochastic nature has triggered these algorithms to be considered in FS [25], [26]. Although the high efficiency of these methods for FS from low-dimensional datasets has been proven, increasing the number of features in a dataset will cause the algorithm to trap in local optimal points and undergo early convergence. The reason is the exponential growth of the problem search space and the addition of many local optimal points in the problem search space [18]. In addition, the evaluation of the solutions obtained by EC methods in a high-dimensional dataset imposes a high computational cost.

This paper, therefore, aims to provide an efficient FS algorithm for high-dimensional datasets to reduce the problems of FS in such datasets. We expect the proposed algorithm to achieve the highest classification accuracy, with less number of features and less computational cost. The main contributions of this paper are summarized as follows:

1. Propose a new FS algorithm called SFE for high-dimensional datasets. The idea of the algorithm is to employ two new operators to achieve the highest classification accuracy with the least number of features and the shortest possible time.
2. Propose a new operator, called the non-selection operator, to perform the exploration phase of the proposed algorithm. The operator explores the search space globally, and finds unimportant features with reasonable computational cost.
3. Propose a new operator, called the selection operator, to perform the exploitation phase of the proposed algorithm. The operator executes a local search on the results obtained by the non-selection operator, and returns the relevant features that have been considered as non-selectable in the exploration phase.
4. Propose a new hybrid algorithm, called SFE-PSO, which is based on the proposed SFE algorithm and an EC method, the particle swarm optimization (PSO) algorithm. The idea behind the proposed hybrid algorithm is to reduce the dimensionality of the datasets through the SFE algorithm, and to use the high potential of the PSO algorithm to search the new low-dimensional dataset for the optimal subset.
5. Propose a general framework, called SFE-ECs, based on the SFE to examine and exhibit the potential of other EC methods for better performance in FS from high-dimensional data.

## II. RELATED WORK

Recent years, researchers have used EC methods to select features in high-dimensional datasets. The main idea of most FS methods based on EC methods involves dividing the extremely large search space into several smaller spaces and apply the evolutionary computation methods to these smaller spaces. The PSO is one of the EC methods that has received the most attention in feature selection from high-dimensional datasets, because of its computational speed, efficiency, ease of implementation, and fewer parameters, as well as proving the efficiency of this algorithm in FS [27]. The current section briefly reviews some of the FS methods proposed before for high-dimensional datasets.

Tran et. al [28] proposed a FS algorithm with variable-length particles based on the PSO algorithm. The idea of employing variable-length particles is to reduce the search space of high-dimensional datasets through particles with a shorter length to reduce computational costs and memory consumption. Using a filter-based FS algorithm, called the symmetric uncertainty (SU) method, the rank of each feature is determined to be used in the initialization of the population. In addition, a mechanism called length changing is proposed to escape the PSO algorithm from local optimal points of the search space. The efficiency of the method is highly dependent on the efficiency of the SU algorithm and the length-changing mechanism.

Song et al. [29] proposed a FS algorithm based on variable-size cooperative co-evolutionary PSO for high-dimensional datasets. In the algorithm, the importance of each feature is determined using a symmetric uncertainty algorithm to reduce the dimensions of the search space in high-dimensional data. After forming smaller subspaces, the multi-swarm PSO algorithm is used to search each subspace. During the search process, various methods are used to remove duplicate particles and add new particles to increase the diversity of the population. Finally, using a crossover method, the solutions obtained by each sub-swarm are combined to obtain the final solution. This algorithm suffers from certain deficiencies, including the high computational cost due to different swarms, and the possibility of trapping in the local optimal points due to the lack of interaction of relevant features under different swarms.

Tan et al. [5] proposed a new method called DimReM to reduce the high dimensionality of the search space using EC methods. This method aims to physically remove some unimportant features from the dataset to reduce the problem search space. Due to the nature of EC methods, if the best member of the population traps in a locally optimal point, other members of the population may also trap in the local optimal. In addition, because reducing the search space size is slow, achieving the expected solution requires a high computational cost.

An evolutionary multitasking-based FS algorithm called PSO-EMT for high-dimensional datasets is proposed in [27]. In the method, two tasks with different search spaces are examined. The first task employs the important features whose importance is determined by the ReliefF algorithm [6], and the second task employs all the features. The algorithm uses a crossover operator to share information between tasks. In addition, two other mechanisms have been used to reduce search space and maintain population diversity. The poor performance of the ReliefF algorithm causes the algorithm to stop at the local optimal points. Also, because all the features are present in the second task, a high computational cost is spent to evaluate the solution obtained by the PSO algorithm. In addition, to use a multi-population optimization algorithm has a high computational cost to perform the search process.

Similarly, an evolutionary multitasking-based FS algorithm and a PSO algorithm for high-dimensional data, called MTPSO, were proposed in [30]. In the algorithm, using a method based on the ReliefF algorithm, large search spaces are converted into several tasks with small search spaces. The multi-population swarm optimization algorithm is then used to search the space of each task. To achieve higher efficiency in the search process using a method, called knowledge transfer, important information about features is exchanged between each subset.

Song et al. [31] presented a FS algorithm called HFS-C-P for high-dimensional datasets. This algorithm consists of three important phases. In the first phase, the symmetric uncertainty algorithm is used to remove unimportant features. In the second phase, the features which are remained in the first phase are clustered using a fast correlation-guided feature clustering method. Finally, in the third phase, the particles of the PSO algorithm are initialized using the features of the clusters, and the search process is performed to obtain an optimal subset. The poor performance of the symmetric uncertainty algorithm in the first phase and the wrong removal of some relevant features can cause the poor performance of the algorithm in certain datasets.

Wang et al. [32] proposed a FS algorithm called SaWDE which is based on the weighted differential evolution algorithm and exploits a self-adaptive mechanism. The multi-population mechanism increases population diversity, and several sub-populations search the problem search space to find the optimal subset of features. Two self-adaptive mechanisms have been used to increase the efficiency of the DE algorithm in the feature selection from different datasets. The mechanisms were proposed to select the appropriate mutation operator and the appropriate value for the parameters of the DE algorithm. In addition, a weight model is used to determine the importance of each feature. The main disadvantage of their proposed method is its high computational cost.

Cheng et al. [33] proposed a steering-matrix-based multi-objective evolutionary algorithm called SM-MOEA for FS of high-dimensional datasets. In the algorithm, the importance of features is calculated using the steering-matrix, which is based on search agents. Therefore, it is employed to guide the population toward better solutions. In addition, in the method, two operators, dimensionality reduction and individual repairing, are used to remove unimportant features from the search agents as well as the search space of the problem. One of the disadvantages of the method is the high memory consumption in creating a steering-matrix from a dataset with a large number of features. Moreover, the wrong removal of relevant and important features using the dimensionality reduction mechanism can cause this algorithm to stop at local optimal points and the performance of the algorithm reduces.

In summary, the above methods aim to reduce the problem search space, to employ EC methods for searching in low dimensional datasets, and to achieve the expected performance in high-dimensional datasets. Nevertheless, some of the above methods, such as [32] and [33], require a lot of memory consumption and high computational cost for feature selection. Some other methods, such as [31], have less computational cost; however, due to the use of filter-based methods for dimension reduction, they may stop at local optimal points in certain datasets. These conditions arise when the relevant features are not determined correctly, and as a result, they are incorrectly filtered from the dataset. Consequently, poor efficiency in classification would be achieved. Overall, the challenges of current feature selection algorithms proposed before can be summarized as follows: in high-dimensional search space, there is the possibility of stopping the algorithm at the local optimal points, the high computational cost and high memory consumption, and poor efficiency in classification. Therefore, this paper proposes an effective and efficient feature selection algorithm for high-dimensional datasets to overcome the above challenges.

## III. THE PROPOSED APPROACH

EC methods have been used as a successful method in FS applications due to their accurate results with reasonable computational cost and memory consumption during the global search [10]. However, while using these methods in high-dimensional datasets, most of them are faced with the curse of dimensionality and the high computational cost [31]. In addition, the exponential growth of the problem search space in high-dimensional dataset causes to produce many local minimal points. Therefore, in these situations, EC methods trap in local optimal points and lead to premature convergence [18]. This paper proposes a FS method to overcome the challenges listed above. The proposed FS exhibits a simple structure, low computational cost, and low memory consumption, and without complex operators. Similar to the EC methods, the proposed method iteratively searches the problem search space. However, the proposed method searches the search space with one search agent, unlike evolutionary algorithms that are population-based.

The high computational cost and high memory consumption of the EC methods in selecting features from high-dimensional datasets is because of storing the search agents in the main memory and then performing the calculation on the solutions obtained by the search agents. Therefore, using one search agent can reduce the computational cost and memory consumption of the proposed FS algorithm.

A FS is a binary problem in which features have two modes: selected and non-selected. In the proposed algorithm, the search agent $X$, which is a solution to the FS problem, displays the modes of each feature. Therefore, agent $X$ is represented as shown in Figure 1.

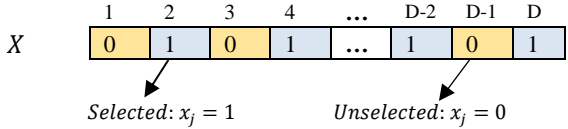

Fig. 1. Encoding of $X$.

where $x_j = 0$ indicates that the $j$th feature is not selected, and $x_j = 1$ indicates that the feature is selected. Therefore, the search agent $X$ is represented using a binary encoding.

The basis of this method is on two simple operators, the selection and the non-selection operators, and their usage on the search agent $X$ in the search process. The selection operator changes the status of a feature from the non-selected mode to the selected mode, and the non-selection operator changes the mode of a feature from the selected mode to the non-selected mode. During the search process of SFE, selection and non-selection operators are applied on the $X$, in order to obtain a better solution of the algorithm. The use of the selection operator depends on the conditions of the problem search space and the position of $X$ in the problem search space. In other words, in the FS process with the SFE algorithm, at each stage of the search process, the non-selection operator is applied on $X$. However, in some datasets, the selection operator may be used from the very beginning of the search process, and in some other datasets, it may never be used until the search process is completed. In this section, we explain the details of the two operators. The pseudo-code of the SFE algorithm is illustrated in Algorithm 1.

### A. The non-selection operator

There are many unimportant features in high-dimensional datasets. On the other hand, FS aims to achieve the highest classification accuracy with the least number of features. Therefore, to achieve this important goal, it is necessary to conduct a global search in the entire search space of the problem to find unimportant features and change their mode to the non-selected mode. To perform a global search in the SFE algorithm, we use the non-selection operator (from line 7 to 12 in Algorithm 1). In any search process, this operator brings the search agent closer to the optimal points of the problem by not selecting a large number of unimportant features. To apply the non-selection operator on $X$, the number of features that the non-selection operator must be applied to is calculated. Different values can be considered for this operator for different datasets. In the SFE algorithm, Equation (1) is used to determine $UN$, which is the number of generated pseudo-random numbers for the operation of the non-selection operator (line 8 in Algorithm 1).

$$UN = [UR \times nvar] \qquad (1)$$

where $nvar$ is the dimensionality of the search space or the number of features of the dataset. The $UR$ is the non-selection operator rate, which can be a number between 0 and 1. If a large value is selected for the $UR$, at each stage of the search process, the status of a large number of features are changed to the non-selected mode by this operator. If the value for the $UR$ is still large, after several stages of the search process, the selection operator would be used in addition to the non-selected operator. Therefore, in such situations, at the beginning of the search process, the non-selection operator is used for a global search, and the algorithm is in the exploration phase. In the following search steps, the algorithm performs the exploitation phase in addition to the exploration phase. A simple method to create a proper balance between the exploration and exploitation phase of the SFE algorithm is to use a linear decrease of the $UR$ value, starting from a large value and gradually changing to a small value. For the SFE algorithm, the $UR$ value is determined using the following equation (line 25 in Algorithm 1):

$$UR = (UR_{max} - UR_{min}) \times ((Max\_FEs - FEs)/FEs) + UR_{min} \qquad (2)$$

where, $UR_{max}$ is the initial value of $UR$, and $UR_{min}$ is the final value of $UR$. $Max\_FEs$ is the maximum number of function evaluations and $FEs$ is the current number of function evaluations of the SFE. After determining the $UN$ value through Equation (1), $UN$ number of pseudo-random number between the 1 and the number of selected features in the solution $X$ is generated (line 10 in Algorithm 1). These pseudo-random numbers are used to access features with selected modes stored in the index vector. In other words, each of the pseudo-random numbers generated is used as an index to the index vector in which the selected feature numbers is stored.

We review the search process of the SFE algorithm on a dataset with 20 features. Figure 2 shows the operation of the non-selection operator through an example. To apply the non-selection operator, the number of features that the non-selection operator must operate on them is calculated using Equation (1). Assuming $UR = 0.3$ and $nvar = 20$, then $UN$ would be 6. This means that six pseudo-random numbers must be generated with a uniform distribution between 1 and 9 (9 is the number of selected features in the $X$). Suppose that the six pseudo-random generated numbers are 3, 7, 9, 7, 3, and 6. These numbers are used to obtain the numbers of the features in the selected mode. As shown in Figure 2, these features are feature numbers 3, 6, 8, 9, 10, 13, 16, 17, and 20. The number of features are stored in the index vector. For this purpose, we obtain the number of features that their mode must be changed to non-selected (line 11 in Algorithm 1). These are feature numbers 8, 16, 20, 16, 8, and 13. According to Figure 2, these features are changed to the non-selected mode, and a new solution called $X_{New}$ is obtained

(line 12 in Algorithm 1). If the fitness value of $X_{New}$ is greater than the fitness value of $X$, then $X_{New}$ will be passed on to the next generation to continue the search process.

---

**Algorithm 1: SFE Feature Selection Algorithm**

**Input:** Training dataset with original feature set, $F = (f_1, f_2, f_3, ..., D)$
**Output:** Training dataset with feature subspace, $S = (s_1, s_2, s_3, ..., d)$

1    $Max\_FEs$: Maximum number of function evaluations
2    $Nvar$: Number of features in dataset
3    Initialize an individual $X = (x_1, x_2, ..., x_D)$ in the search space
4    Calculate the fitness of $X$: $fit(X)$
5    $FEs = 0$
6    **While** $FEs \leq Max\_FEs$ **Do**
7       $X_{New} = X$
8       $UN = [UR \times nvar]$          % The number of features to change to non-selected mode
9       $Index$=find the indexes of selected features in $X$
10      $U$=Generate $UN$ random number between 1 to the number of selected features in $X$
11      $K = index(U)$
12      Set $X_{New}(K) = 0$          **% non-selection operation**
13      **If** the number of selected features in $X_{New}$ is $== 0$
14         $X_{New} = X$
15         $Index$=find the indexes of selected features in $X$
16         $SN = 1$          % The number of features to change to select mode
17         $S$=Generate $SN$ random number between 1 to the number of unselected features in $X$
18         $K = index(S)$
19         Set $X_{New}(K) = 1$          **% selection operation**
20      **End if**
21      **If** $fit(X_{New}) \geq fit(X)$
22         $X = X_{New}$
23      **End if**
24      $FEs = FEs + 1$
25      $UR = (UR_{max} - UR_{min}) \times ((Max\_FEs - FEs)/FEs) + UR_{min}$
26    **End while**

---

Fig. 2. Illustration of non-selection operation.

Now suppose the $X_{New}$ fitness value is greater than the $X$ fitness value, so the search process continues using the $X_{New}$. Assuming $UR = 0.1$, then the $UN$ value would be 2. This means that two pseudo-random numbers with a uniform distribution must be generated between the number 1 and 5 (5 is the number of selected features in the $X$). Suppose that two pseudo-random numbers are 3 and 5. In this case, among the features selected in $X$, that is, feature numbers 9, 6, 3, 10, and 17, the two features must be changed to the non-selected mode by the non-selection operator. For this purpose, by the command $K = index(U)$, their numbers would be obtained, 3 and 17. In this way, these two features are changed to the non-selected mode using the non-selection operator. After this process, if the fitness value of $X_{New}$ is greater than $X$, the $X_{New}$ will be passed on to the next generation to continue the search process. But if the fitness value of $X_{New}$ is not greater than the fitness value of $X$, the search process continues with the $X$.

### B. The selection operator

After performing a global search using the non-selection operator, all the features in the dataset may be changed to the non-selected mode. In such situations, the selection operator would be applied (from line 13 to 20 in Algorithm 1). The selection operator allows the algorithm to re-select important features whose status has changed to non-selected mode. Assume that we perform one more step with the non-selection operator, and the $X$ is changed to the $X_{New}$, as shown in Figure 3. All the features in the $X_{New}$ solution have been changed to non-selected mode, and the algorithm does not change the $X$ solution for a better subset in its global search. Therefore, the selection operator is used. To do this, the $X_{New}$ is first changed to the $X$ according to the Figure 3, and $SN$ number of pseudo-random values is generated between the 1 and the number of features that are not selected. Assuming that $SN = 1$ and the pseudo-random number generated is 13, then the feature number 16 changes to the selected mode. As FS aims to achieve the highest accuracy with the least number of features, a small value should usually choose for the $SN$. On the other hand, because the algorithm may choose a small number of features after several search steps, using large values for the $SN$ may not achieve the desired improvement. Therefore, a small value such as 1 can be suitable for $SN$. This operator makes small changes in solution $X$, and its behavior is similar to performing a local search with minimal changes around the best solutions; this operator increases the exploitation capability of the SFE algorithm, consequently.

Fig. 3. Illustration of selection operation.

### C. Fitness Function

The classification accuracy is used to evaluate the new solutions obtained at each step of the search process as follow:

$$fit = \frac{T_P + T_N}{T_P + T_N + F_P + F_N} \times 100\% \qquad (3)$$

where $T_P$ and $T_N$ are the number of positive and negative instances of the dataset, which a classifier has correctly classified as a member of the positive and negative class members. Also, $F_P$ and $F_N$ are the number of positive and negative instances of the dataset that a classifier has incorrectly classified as negative and positive class members.

(a)  The training dataset before feature selection

(b)  The training dataset after feature selection using SFE algorithm

(c) The training dataset after reducing the search space by removing unselected features from dataset

Fig. 4. Illustration of proposed hybrid algorithm.

### D. The hybridization of SFE and PSO

Although EC methods are not very efficient in FS from high-dimensional datasets, the high efficiency of these methods in FS in low-dimensional datasets has been proven [34]-[40]. Therefore, if the problem search space in high-dimensional dataset can be reduced by physically removing unimportant features from the dataset, the high potential of EC methods can be used to achieve better results.

This paper uses this idea to propose a hybrid algorithm based on the SFE and the PSO algorithm. Figure 4 shows the process of the proposed hybrid algorithm. The proposed hybrid algorithm consists of three steps as follows:

1. To use the SFE algorithm in the early stages of the search by conducting a global search throughout the large search space.

2. To reduce the dimensionality of the search space by physically removing non-selected mode features after completing the SFE algorithm search steps.

3. To use PSO algorithm to search the dimensionality reduced search space.

In a dataset, all features are in selected mode before performing the feature selection (Figure 4(a)). First, the SFE algorithm is used to select important features and to not select unimportant features from the dataset (Figure 4(b)). After selecting the relevant features with the SFE algorithm, all non-selected features are physically removed from the dataset, and a new dataset is created with dimensionality much smaller than the original dataset (Figure 4(c)). Therefore, the PSO algorithm can obtain an optimal feature subset in this new search space than the subset obtained by SFE. In Algorithm 2, the pseudo-code of the SFE-PSO hybrid algorithm is presented.

---

**Algorithm 2: Proposed SFE-PSO Algorithm**

|   | |
|---|---|
|   | **Input:** Training dataset with original feature set, $F = (f_1, f_2, f_3, \ldots, D)$ |
|   | **Output:** Training dataset with feature subspace, $S = (s_1, s_2, s_3, \ldots, d)$ |
| 1 | $Max\_FEs$: Maximum number of function evaluations |
| 3 | Initialize an individual $X = (x_1, x_2, \ldots, x_D)$ in the search space |
| 4 | Calculate the fitness of $X$: $fit(X)$ |
| 5 | $FEs = 0$ |
| 6 | **While** $FEs \leq Max\_FEs$ **Do** |
| 7 |     Select and unselect features using **Algorithm 1** and save the fitness of the best solution in the fitness vector |
| 8 |     **If** $FEs > 2000$ and $(\text{fitness vector}(FEs) - \text{fitness vector}(FEs - 1000) == 0)$ |
| 9 |         S = removing unselected features from the training dataset |
| 10 |         Initialize particles with positions x and velocity v. |
| 11 |         $x_i = (x_{i,1}, x_{i,2}, \ldots, x_{i,d}), \quad v_i = (v_{i,1}, v_{i,2}, \ldots, v_{i,d}), \; i = 1,2,\ldots,N_p, \; j = 1,2,\ldots,d$ |
| 12 |         Set $x_{1,j=1\ldots d} = 1$, |
| 13 |         Evaluate the fitness for each particle $fit(x_1), \ldots, fit(x_{N_p})$ |
| 14 |         Update pbest and gbest for each particle |
| 15 |         **While** $FEs \leq Max\_FEs$ **Do** |
| 16 |             **For** $i = 1$ to $N_p$ **Do** |
| 17 |                 **For** $j = 1$ to $d$ **Do** |
| 18 |                     $v_{i,j} = w\,v_{i,j} + c_1 r_1 (pbest_{i,j} - x_{i,j}) + c_2 r_2 (gbest_{i,j} - x_{i,j})$ |
| 19 |                     $x_{i,j} = \begin{cases} 1, & if\ rand() \leq Z(v_{i,j}) \\ 0, & Otherwise \end{cases}$ |
| 20 |                 **End for** |
| 21 |                 Evaluate the fitness for each particle $fit(x_1), \ldots, fit(x_{N_p})$ |
| 22 |                 Update gbest and pbest for each particle |
| 23 |             **End for** |
| 24 |             $FEs = FEs + N_p$ |
| 25 |         **End while** |
| 26 |     **End if** |
| 27 |     $FEs = FEs + 1$ |
| 28 | **End while** |

---

As we can see in the pseudo-code of Algorithm 2, first, the SFE algorithm is used to select the relevant features (line 7 in Algorithm 2). Then, suppose the function evaluation counter is greater than 2000, and the SFE algorithm has not been able to get a better solution in its previous 1000 function evaluations. In that case, the search process continues under the new S space with the PSO algorithm (line 8 in Algorithm 2).

To continue the search process with the PSO algorithm in the new subspace, first, the features in non-selected mode are physically removed from the original dataset and a new dataset with lower dimensionality is created. Then, after initializing the particle velocity and position values, the position of the first particle is changed to the best solution obtained by the SFE algorithm. It means that, in the first particle, all the features are in the selected mode. In this way, the PSO algorithm continues the process for better solutions in the new space, which is a much smaller space than the original search space.

### E. ECs-Based Feature Selection with SFE

To achieve better solution for the FS problem, we can combine the SFE algorithm with various EC methods. In Algorithm 3, the general framework of the SFE-ECs hybrid algorithm is illustrated. The SFE-ECs consist of two phases. In the first phase, the SFE algorithm is used to select the feature in the huge search space of high-dimensional data (line 6 in Algorithm 3). Non-selected features are then physically removed from the original feature set (line 8 in Algorithm 3). In the second phase, the search process continues with EC methods in the new search space, which is a much smaller search space than the original ones (line 9 in Algorithm 3).

Overall, the FS process starts using the SFE algorithm, and this process continues until the specified condition for completing the search process of the SFE algorithm is met. Then the search process continues with the EC methods proposed for FS in low-dimensional datasets such as PSO [35], [36], DE [37], GA [38], Grey Wolf Optimization (GWO) [39], Salp Swarm Algorithm (SSA) [40], Ant Colony Optimization (ACO) [41], and the Artificial Bee Colony (ABC) algorithm [42].

## IV. EXPERIMENT DESIGN

To evaluate the effectiveness and efficiency of the proposed FS methods, experiments have been performed on 40 real datasets. Nine algorithms have been used to compare with the proposed FS methods. First, the results of SFE and SFE-PSO algorithms are compared with the six algorithms BDE [43], BPSO [44], BGA [5], DimReM-BDE [5], DimReM-BGA [5], and DimReM-PSO [5]. Then, the results of the proposed algorithms are compared with more three state-of-the-art EC-based feature selection algorithms, including, HFS-C-P [31], SaWDE [32], and SM-MOEA [33] algorithms. The reason for choosing the three algorithms BDE, BPSO, and BGA, is to evaluate the performance of these algorithms as three well-known EC algorithms in FS on high-dimensional datasets. Also, due to the high efficiency of the three algorithms, DimReM-BDE, DimReM-BGA, and DimReM-PSO, in reducing the dimensionality of data, these algorithms also compared with the proposed method. The three algorithms HFS-C-P, SaWDE, and SM-MOEA are also chosen since they were recently proposed for FS in the high-dimensional datasets.

**Algorithm 3: Framework of SFE-EC**

| | |
|---|---|
| | **Input:** Training dataset with original feature set, $F = (f_1, f_2, f_3, \ldots, D)$ |
| | **Output:** Training dataset with feature subspace, $S = (s_1, s_2, s_3, \ldots, l)$ |
| 1 | $Max\_FEs$: Maximum number of function evaluations |
| 3 | Initialize an individual $X = (x_1, x_2, \ldots, x_D)$ in the search space |
| 4 | Calculate the fitness of $X$: $fit(X)$ |
| 5 | **While** the stopping criterion is not met **Do** |
| 6 |     *Select and unselect features using **Algorithm 1*** |
| 7 |     **If** the stopping criterion is met |
| 8 |         S = removing the unselected features from the training dataset |
| 9 |         **Call EC methods for feature selection in new subspace, $\mathbf{S = (s_1, s_2, s_3, \ldots, l)}$** |
| 10 |     **End if** |
| 11 | **End while** |

More information about experiments is reported in section I, in the supplementary material. The source codes of the proposed algorithms are publicly available at https://github.com/Ahadzadeh2022/SFE.git.

## V. RESULTS AND DISCUSSIONS

### A. Classification accuracy

The classification accuracy is one of the important criteria for evaluating the effectiveness of an FS algorithm. In this section, the performance of the proposed algorithms is compared with that of other algorithms based on the classification accuracy. Wilcoxon rank-sum test with a significance level of 0.05 is adopted to evaluate the performance of the SFE-PSO with 30 independent runs. In the following results, "+", "−", and "≈" indicate that the SFE-PSO is significantly better than, worse than, and similar to the compared algorithms. Also, the mean ranking of the Friedman test is utilized to compare the effectiveness of the SFE-PSO algorithm with other algorithms and detect significant improvements. Table I illustrates the results of the algorithms over the 30 independent runs.

As illustrated in Table I, the average classification accuracy of the SFE algorithm is better than that of other six algorithms in the 30 independent runs, almost in most datasets examined. The superiority of the SFE algorithm over other algorithms in some datasets is significantly high. For example, in the warpAR10P and prostate cancer datasets, the SFE algorithm outperformed the best algorithms of the other six algorithms by 25.17% and 23.40% better accuracy, respectively. However, this improvement is very small in some datasets. For example, in the Brain_Tumor1 dataset, the SFE algorithm is 0.89% superior to the BPSO algorithm, which is the best algorithm among the other six algorithms.

Examining and comparing the results of the SFE algorithm and SFE-PSO algorithm shows the superiority of the SFE-PSO algorithm in 22 out of 40 datasets. This improvement is significantly high in some datasets. For example, in the CML treatment and Carcinom datasets, the SFE-PSO algorithm is 6.16% and 5.60% superior to the SFE algorithm, respectively. This improvement is very small in some datasets. For example, in the SMK_CAN_187 datasets, the SFE-PSO algorithm is 0.0981% superior to the SFE algorithm. Figure 5 shows the results of the algorithms at the average classification accuracy in 40 datasets. The BPSO algorithm, with an average classification accuracy of 83.15 %, is the best algorithm among the six algorithms: BDE, BPSO, BGA DimReM-BDE, DimReM-BGA, and DimReM-BPSO. Comparing the efficiency of SFE and SFE-PSO algorithms with that of the BPSO algorithm indicates the superiority of 8.71% and 10.13% of these algorithms over the BPSO algorithm, which is significantly high. The Wilcoxon and Friedman test results have demonstrated the superiority of the SFE-PSO algorithm since it is significantly better than or similar to other algorithms in all datasets in classification accuracy criteria.

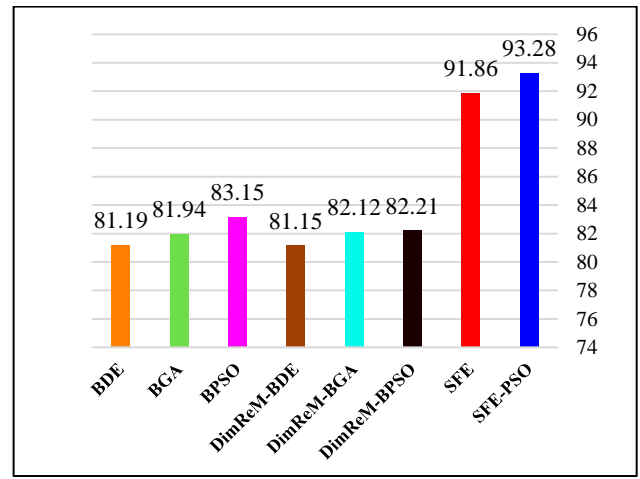

Fig. 5. Average classification accuracy by the eight algorithms.

Figure 6 shows the convergence diagram of the SFE algorithm, the SFE-PSO algorithm, and the other six algorithms, which have important information about the behavior of the algorithms in the FS process. All six algorithms, BDE, BGA, BPSO, DimReM-BDE, DimReM-BGA, and DimReM-BPSO, have fixed behavior in all 40 datasets. These algorithms start their search with a random point in the problem search space and get better solutions in the early stages of the search, and finally, they stop in a huge search space with a large number of local optimal points. The relative superiority of the BPSO algorithm over other algorithms can be seen. In contrast, observing the behavior of the SFE, SFE-PSO algorithms is very significant. In most datasets such as the Breast Cancer, DLBCL, and GLI_85 datasets, the SFE has an expected behavior: starting the search process with a random point and achieving better solutions during the search process. More information about Figure 6 is reported in Section III, in the supplementary material. Tables III and IV in the supplementary material present the results of the algorithms in the best, worst, mean, and standard deviation of classification accuracy in 30 independent runs on 40 datasets.

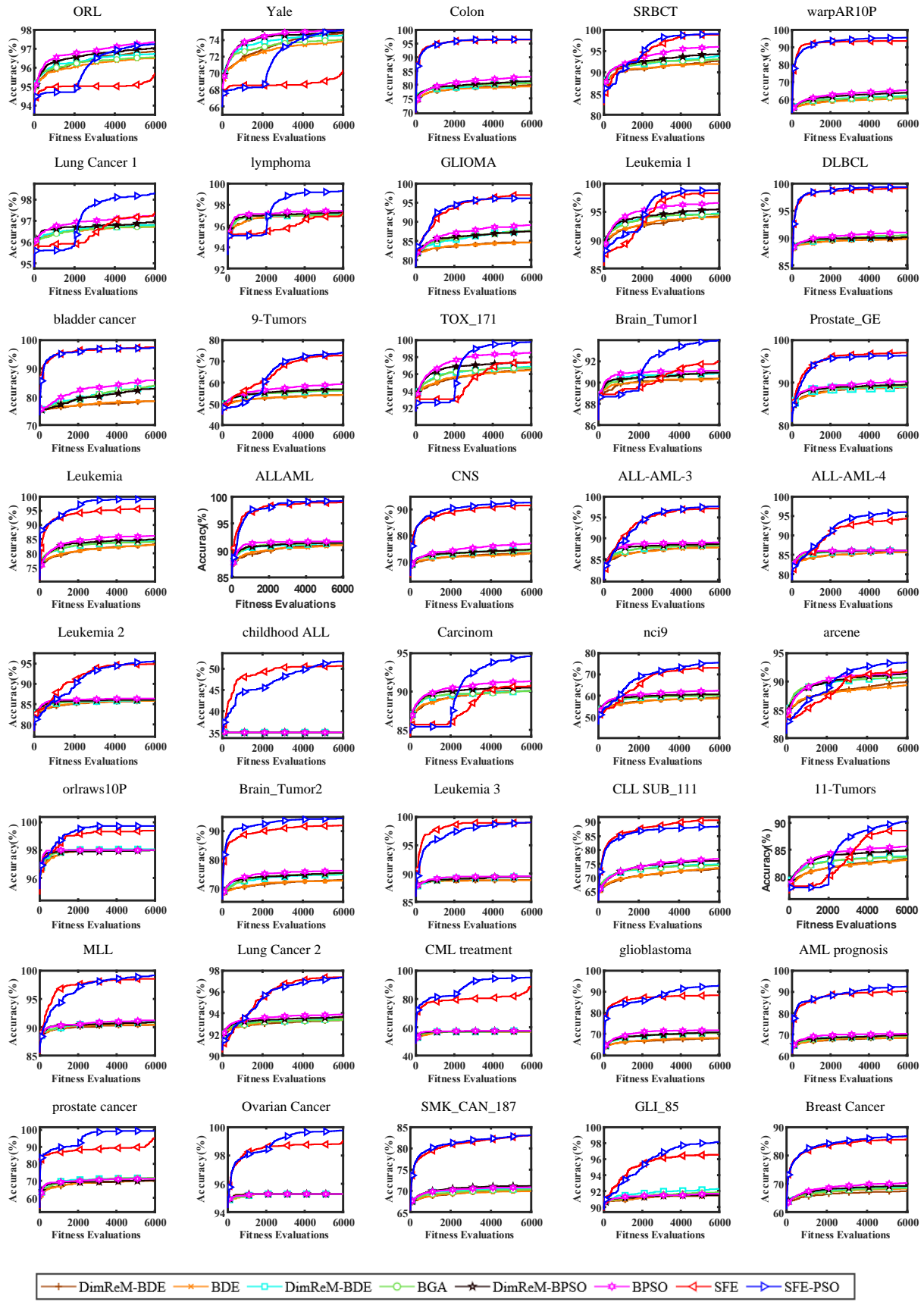

Fig. 6. Convergence curve of SFE and SFE-PSO versus other algorithms over all datasets - KNN classifier.

TABLE I
THE MEAN AND STD OF THE CLASSIFICATION ACCURACY OBTAINED BY EIGHT ALGORITHMS OVER THE 30 INDEPENDENT RUNS

| No. | Dataset | BDE [43] | BGA [5] | BPSO [44] | DimReM-BDE [5] | DimReM-GA [5] | DimReM-BPSO [5] | SFE | SFE-PSO |
|---|---|---|---|---|---|---|---|---|---|
| 1 | ORL | 96.49±0.18 (+) | 96.55±0.27 (+) | **97.35±0.22** (+) | 96.86±0.19 (+) | 96.73±0.22 (+) | 97.05±0.31 (≈) | 95.71±0.66 (+) | 97.24±0.57 |
| 2 | Yale | 73.82±0.40 (+) | 74.02±0.73 (+) | 75.15±0.26 (≈) | 74.82±0.36 (≈) | 74.57±0.51 (≈) | 74.95±0.39 (≈) | 70.41±1.20 (+) | **75.23±1.35** |
| 3 | Colon | 79.30±0.91 (+) | 80.46±1.19 (+) | 82.84±1.20 (+) | 79.90±0.99 (+) | 81.23±1.59 (+) | 81.34±1.55 (+) | 96.44±2.44 (≈) | **96.51±2.46** |
| 4 | SRBCT | 91.94±0.72 (+) | 93.04±1.42 (+) | 97.64±0.79 (+) | 92.66±0.64 (+) | 93.84±1.60 (+) | 94.27±1.20 (+) | 98.89±1.99 (≈) | **99.09±1.14** |
| 5 | warpAR10P | 60.56±0.72 (+) | 63.84±1.10 (+) | 68.46±1.25 (+) | 61.89±0.69 (+) | 65.38±1.18 (+) | 63.87±1.28 (+) | 93.64±3.39 (+) | **95.53±2.29** |
| 6 | Lung Cancer 1 | 96.70±0.22 (+) | 96.73±0.29 (+) | 97.24±0.33 (+) | 96.81±0.24 (+) | 96.80±0.38 (+) | 96.93±0.36 (+) | 97.41±0.64 (≈) | **98.32±0.37** |
| 7 | lymphoma | 97.00±0.31 (+) | 97.03±0.36 (+) | 97.42±0.60 (+) | 97.11±0.42 (+) | 97.14±0.45 (+) | 97.28±0.95 (+) | 97.28±0.95 (+) | **99.35±0.95** |
| 8 | GLIOMA | 84.53±1.04 (+) | 87.73±1.55 (+) | 89.13±1.71 (+) | 84.66±0.95 (+) | 87.66±1.66 (+) | 87.53±2.01 (+) | **97.00±2.61** (≈) | 96.13±2.09 |
| 9 | Leukemia 1 | 94.06±0.82 (+) | 94.60±1.08 (+) | 96.57±0.68 (+) | 94.25±0.83 (+) | 94.74±0.97 (+) | 95.49±0.97 (+) | 98.28±1.38 (+) | **98.87±0.72** |
| 10 | DLBCL | 86.79±0.03 (+) | 90.46±0.86 (+) | 91.05±0.77 (+) | 89.91±0.63 (+) | 90.50±0.78 (+) | 90.17±1.00 (+) | 99.16±1.10 (≈) | **99.36±1.03** |
| 11 | bladder cancer | 78.40±2.48 (+) | 84.00±3.95 (+) | 85.90±4.83 (+) | 78.40±2.02 (+) | 83.60±4.62 (+) | 82.80±5.46 (+) | **97.50±2.16** (≈) | 97.30±2.69 |
| 12 | 9-Tumors | 54.22±1.36 (+) | 56.22±1.37 (+) | 59.22±1.39 (+) | 54.16±0.95 (+) | 56.27±1.43 (+) | 56.88±1.79 (+) | 72.77±7.80 (+) | **74.16±4.92** |
| 13 | TOX_171 | 96.44±0.40 (+) | 96.73±0.54 (+) | 98.54±0.42 (+) | 96.54±0.46 (+) | 96.87±0.61 (+) | 97.37±0.52 (+) | 97.44±1.50 (+) | **99.78±0.32** |
| 14 | Brain_Tumor1 | 90.37±0.53 (+) | 90.96±0.38 (+) | 91.11±2.8e-14 (+) | 90.37±0.53 (+) | 90.85±0.47 (+) | 90.92±0.42 (+) | 92.00±1.14 (+) | **94.00±0.95** |
| 15 | Prostate GE | 88.90±0.50 (+) | 89.09±0.61 (+) | 90.30±0.63 (+) | 88.86±0.51 (+) | 88.92±0.66 (+) | 89.51±0.78 (+) | **97.05±1.45** (≈) | 96.34±1.68 |
| 16 | leukemia | 82.92±0.90 (+) | 84.00±1.15 (+) | 87.42±0.68 (+) | 83.21±0.67 (+) | 84.28±1.31 (+) | 84.91±1.16 (+) | 95.82±2.03 (+) | **99.05±0.72** |
| 17 | ALLAML | 90.90±0.68 (+) | 91.42±0.49 (+) | 91.61±4.3e-14 (+) | 91.16±0.64 (+) | 91.33±0.67 (+) | 91.33±0.57 (+) | 98.93±0.14 (≈) | **99.22±1.11** |
| 18 | CNS | 73.00±1.10 (+) | 74.22±1.04 (+) | 77.05±2.08 (+) | 73.27±0.92 (+) | 74.22±1.21 (+) | 74.61±1.67 (+) | 91.50±3.34 (≈) | **92.61±2.79** |
| 19 | ALL-AML-3 | 87.76±0.61 (+) | 88.33±0.65 (+) | 89.00±0.54 (+) | 88.17±0.67 (+) | 88.33±0.65 (+) | 88.57±0.60 (+) | 97.13±2.18 (≈) | **97.68±1.55** |
| 20 | ALL-AML-4 | 85.68±0.57 (+) | 86.09±0.36 (+) | 86.09±0.36 (+) | 85.68±0.57 (+) | 86.04±0.26 (+) | 86.09±0.36 (+) | 94.40±3.10 (+) | **96.00±1.76** |
| 21 | Leukemia 2 | 85.82±0.46 (+) | 85.95±0.24 (+) | 86.38±0.64 (+) | 85.86±0.54 (+) | 86.00±0.00 (+) | 86.04±0.26 (+) | 94.87±2.34 (+) | **95.60±1.70** |
| 22 | childhood ALL | 35.00±0.00 (+) | 35.00±0.00 (+) | 35.00±0.00 (+) | 35.00±0.00 (+) | 35.00±0.00 (+) | 35.00±0.00 (+) | 50.66±17.1 (≈) | **51.80±15.7** |
| 23 | Carcinom | 90.08±0.33 (+) | 90.11±0.59 (+) | 91.36±0.45 (+) | 90.08±0.42 (+) | 90.08±0.60 (+) | 90.50±0.58 (+) | 90.86±2.25 (+) | **96.47±0.98** |
| 24 | nci9 | 58.72±0.94 (+) | 61.66±1.10 (+) | 62.38±1.13 (+) | 59.11±0.95 (+) | 60.11±1.23 (+) | 60.61±1.11 (+) | 73.11±4.62 (+) | **75.55±3.31** |
| 25 | arcene | 89.36±0.69 (+) | 90.66±1.20 (+) | 91.69±0.66 (+) | 89.90±0.78 (+) | 90.71±0.82 (+) | 91.40±0.67 (+) | 92.01±3.24 (+) | **93.36±1.75** |
| 26 | orlraws10P | 98.00±0.00 (+) | 98.00±0.00 (+) | 98.00±0.00 (+) | 98.03±0.18 (+) | 98.06±0.25 (+) | 98.00±0.37 (+) | 99.40±0.77 (≈) | **99.73±0.44** |
| 27 | Brain_Tumor2 | 72.46±1.45 (+) | 75.00±1.14 (+) | 76.00±0.90 (+) | 72.73±1.11 (+) | 74.66±1.09 (+) | 75.20±1.62 (+) | 92.00±3.36 (+) | **94.46±2.90** |
| 28 | Leukemia 3 | 88.82±0.38 (+) | 89.46±0.92 (+) | 89.52±0.92 (+) | 88.87±0.46 (+) | 89.52±0.71 (+) | 89.41±1.00 (+) | 99.00±1.29 (≈) | **99.06±1.70** |
| 29 | CLL_SUB_111 | 73.19±0.78 (+) | 74.47±1.27 (+) | 76.79±1.01 (+) | 73.48±0.77 (+) | 77.39±1.47 (+) | 76.04±1.06 (+) | **90.81±3.07** (≈) | 88.59±2.93 |
| 30 | 11-Tumors | 83.06±0.57 (+) | 83.59±0.93 (+) | 85.67±0.52 (+) | 83.31±0.62 (+) | 83.77±0.75 (+) | 84.85±0.73 (+) | 88.54±2.09 (+) | **90.29±1.24** |
| 31 | MLL | 90.41±0.50 (+) | 90.81±0.77 (+) | 91.21±0.76 (+) | 90.41±0.61 (+) | 91.04±0.75 (+) | 90.90±0.68 (+) | 98.55±1.54 (≈) | **99.17±1.27** |
| 32 | Lung Cancer 2 | 93.48±0.27 (+) | 93.36±0.38 (+) | 94.09±0.28 (+) | 93.32±0.28 (+) | 93.40±0.32 (+) | 93.64±0.40 (+) | **97.38±0.71** (≈) | 97.37±0.76 |
| 33 | CML treatment | 56.66±1.4e-14 (+) | 56.96±1.01 (+) | 57.46±1.63 (+) | 57.46±1.63 (+) | 57.44±1.49 (+) | 57.46±1.63 (+) | 89.01±5.94 (+) | **95.17±3.87** |
| 34 | glioblastoma | 78.08±1.35 (+) | 70.32±1.37 (+) | 71.68±0.74 (+) | 67.84±0.98 (+) | 70.64±1.25 (+) | 70.64±1.38 (+) | 91.50±3.34 (+) | **92.96±3.06** |
| 35 | AML prognosis | 68.29±0.82 (+) | 69.09±0.90 (+) | 70.25±0.82 (+) | 68.36±1.04 (+) | 69.30±1.30 (+) | 69.60±1.01 (+) | 88.32±3.85 (+) | **92.58±4.05** |
| 36 | prostate cancer | 70.00±0.00 (+) | 71.80±2.44 (+) | 71.40±2.70 (+) | 70.20±1.00 (+) | 71.60±2.38 (+) | 70.20±2.27 (+) | 95.20±4.89 (+) | **99.40±1.65** |
| 37 | Ovarian Cancer | 95.27±1.4e-14 (+) | 95.27±1.4e-14 (+) | 95.27±1.4e-14 (+) | 95.27±1.4e-14 (+) | 95.27±1.4e-14 (+) | 95.27±1.4e-14 (+) | 99.10±0.70 (≈) | **99.76±0.42** |
| 38 | SMK_CAN_187 | 69.88±0.44 (+) | 70.19±0.52 (+) | 70.69±0.46 (+) | 70.08±0.43 (+) | 70.54±0.57 (+) | 71.16±0.53 (+) | 83.03±2.36 (≈) | **83.13±3.18** |
| 39 | GLI_85 | 91.52±0.47 (+) | 91.85±0.75 (+) | 91.67±0.67 (+) | 91.45±0.52 (+) | 92.27±0.66 (+) | 91.48±0.70 (+) | 96.54±1.27 (+) | **98.15±1.23** |
| 40 | Breast Cancer | 66.86±0.84 (+) | 68.45±0.45 (+) | 70.30±0.77 (+) | 66.86±0.84 (+) | 68.78±0.75 (+) | 69.21±1.21 (+) | 85.70±4.15 (+) | **86.83±3.97** |
| | Wilcoxon Test (+ \| ≈ \| −) | 40 \| 0 \| 0 | 40 \| 0 \| 0 | 38 \| 2 \| 0 | 39 \| 1 \| 0 | 39 \| 1 \| 0 | 38 \| 2 \| 0 | 22 \| 17 \| 1 | – |
| | Friedman Test (Mean Rank) | 7.40 | 5.53 | 3.29 | 6.79 | 5.01 | 4.58 | 2.26 | **1.15** |

## B. Number of selected features

The number of selected features is another important criterion to evaluate the effectiveness of the FS algorithms. Table V in the supplementary material shows the results of the algorithms in terms of the average number of selected features in 30 independent runs. The SFE algorithm has selected less than 100 features from more than thousands of features in 33 datasets out of 40 datasets, which is significant compared to the selected features by the other six algorithms. For instance, out of 12,627 prostate cancer features, the SFE algorithm has selected 24 features. The other six algorithms have selected about half of the features from each dataset.

In addition, the results of the SFE-PSO algorithm are similar to those of the SFE. The notable point in comparing the SFE algorithm and the SFE-PSO algorithm is that in some datasets such as Leukemia 1, TOX_171, Brain_Tumor1, and Lymphoma, the number of selected features of the SFE-PSO algorithm is more than the SFE algorithm. In general, when the SFE algorithm in 2000 evaluation specified in the SFE-PSO algorithm for reasons such as trapping into a local minimum cannot achieve a smaller number of features and improve the classification accuracy, then the search in the new space continues with the PSO algorithm, and according to the properties of the PSO algorithm, which does not have many capabilities in not selecting a large number of features, and the search space conditions of the problem average about half of the features change to non-selected mode. These conditions are well illustrated in Figure 6, which shows the convergence diagrams of the SFE algorithms and the SFE-PSO algorithm in Leukemia 1, TOX_171, Brain_Tumor1, and Lymphoma datasets. Observing the convergence diagram of these algorithms in the mentioned datasets in Figure 6 shows that the SFE algorithm did not succeed in the first 2000 function evaluations; therefore, the number of selected features of the algorithm is much more than the number of features that the algorithm finds in the next 4000 function evaluations. When the search process continues with the PSO algorithm after the first 2000 function evaluations, this algorithm changes about half of the features from selected to non-selected mode. Therefore, in a dataset that the number of selected features of the SFE-PSO is more than the SFE, the PSO algorithm continues the search process in the new space by about twice the number of features found in the final solution. In addition, the table shows that in some other datasets, such as Ovarian Cancer and DLBCL, the number of selected features of the SFE-PSO algorithm is less than the SFE algorithm. This case indicates that the SFE algorithm was unsuccessful in decreasing the number of features and increasing the classification accuracy in the next 4000 evaluations; however, in the continuation with PSO, better results were achieved with fewer features.

Figure 7 shows the average number of selected features by the eight algorithms on the 40 datasets. As we can see, the number of selected features of SFE and SFE-PSO algorithms is 92 and 121, respectively, which are significantly less than the selected features of other algorithms, which are more than 4400 features on average. The analysis of the running time of the proposed algorithms is reported in section II of the supplementary material.

To further evaluate the effectiveness and efficiency of the proposed algorithms, the results of these methods are compared with three state-of-the-art EC-based FS algorithms, namely, HFS-C-P [31], SaWDE [32] and SM-MOEA [33] algorithms. Experiments have been carried out in two parts. The effectiveness and efficiency of the proposed algorithms have been investigated in comparison with the SaWDE and SM-MOEA algorithms on 11 different datasets used in SM-MOEA [33]. In these experiments, the results of SaWDE and SM-MOEA algorithms are performed according to the previous section. That is, in the FS process, the KNN classifier with K=1 has been used in the evaluation function. Furthermore, 5-fold cross-validation method is used to divide the dataset into train and test sets. Each algorithm has been run 30 times independently. The maximum number of fitness function evaluations is set to 6000. Other parameters of both algorithms are the parameters suggested in the algorithms. In the second part, the results of the proposed algorithms are compared with the results of the HFS-C-P algorithm on 8 datasets that were used in the experiments of both papers. Therefore, the results of SFE and SFE-PSO algorithms in this section are based on the fitness function (SVM and KNN classifier whit K=3 and 10-fold cross-validation) and the maximum number of function evaluations (5000 times) considered in HFS-C-P [31]. Tables II show the results of the proposed algorithms, SaWDE and SM-MOEA algorithms on 11 datasets used in SM-MOEA [33] in terms of the best, mean, and standard deviation (Std) of the classification accuracies, the number of selected features(size) and running time (average of the running time in 30 independent runs).

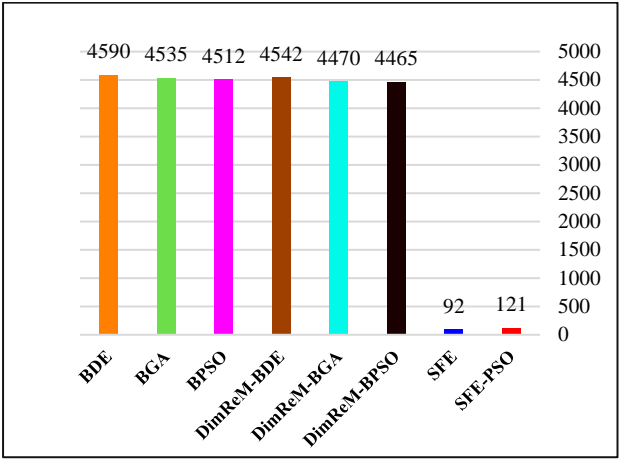

Fig. 7. The average number of selected features by the eight algorithms.

As can be seen from Table II, the mean classification accuracy of SFE-PSO is higher than that of the SM-MOEA algorithm on 8 out of 11 datasets. In the other three datasets, the results of both methods are similar. The Wilcoxon and Friedman test results have also demonstrated the superiority of SFE-PSO, since it is always significantly better or similar to SM-MOEA algorithms. In addition, the table results show that the mean classification accuracy of SFE-PSO is higher than that of the SaWDE algorithm on 4 out of 11 datasets. The results of the two algorithms in six datasets are similar, and in one dataset, the SaWDE algorithm is better than the SFE-PSO algorithm. In general, the results of the Wilcoxon and Friedman tests show the superiority of the SFE-PSO algorithm over the SaWDE algorithm.

TABLE II
COMPARED RESULTS OF SFE, SFE-PSO, SaWDE [32] and SM-MEOA [33]

| No. | Dataset | Algorithm | Size | Time | Best | Mean±Std |
|---|---|---|---|---|---|---|
| 1 | Lung Cancer 1 | SM-MEOA | **38.267**(−) | 463 (+) | **99.02** | 97.89±0.4(≈) |
| | | SaWDE | 149.23(+) | 1343 (+) | 98.53 | 98.09±0.1(≈) |
| | | SFE | 127.4 (+) | **262** (−) | 98.53 | 97.41±0.6(≈) |
| | | SFE-PSO | 77.9 | 281 | **99.02** | **98.32±0.3** |
| 2 | GLIOMA | SM-MEOA | **18.76** (−) | 503 (+) | **100** | 94.53±2.1(+) |
| | | SaWDE | 87.72 (+) | 1710 (+) | **100** | 95.60±2.1(≈) |
| | | SFE | 26.8 (−) | **235** (−) | **100** | **97.00±2.6**(≈) |
| | | SFE-PSO | 28.633 | 249 | **100** | 96.13±2.0 |
| 3 | Leukemia 1 | SM-MEOA | 59.36 (−) | 560 (+) | **100** | 98.25±1.0(+) |
| | | SaWDE | 212 (+) | 2795 (+) | **100** | 98.49±0.5(≈) |
| | | SFE | **46.43** (+) | 250 (−) | **100** | 98.28±1.3(+) |
| | | SFE-PSO | 115.5 | 261 | **100** | **98.78±0.7** |
| 4 | DLBCL | SM-MEOA | **18.06** (≈) | 592 (+) | **100** | 99.23±0.9(≈) |
| | | SaWDE | 188 (+) | 3106 (+) | **100** | 97.79±1.4(+) |
| | | SFE | 29 (+) | 244 (−) | **100** | 99.16±1.1(≈) |
| | | SFE-PSO | 18.93 | 248 | **100** | **99.36±1.0** |
| 5 | 9-Tumors | SM-MEOA | **36.6** (−) | 642 (+) | 76.66 | 65.55±5.2(+) |
| | | SaWDE | 301.65(+) | 2573 (+) | 82.84 | **74.32±7.8**(≈) |
| | | SFE | 65.83 (−) | **250** (−) | **86.66** | 72.77±7.8(+) |
| | | SFE-PSO | 152.63 | 265 | 85.00 | 74.16±4.9 |
| 6 | Brain Tumor 1 | SM-MEOA | **67.40** (+) | 689 (+) | 95.55 | 92.62±1.3 (+) |
| | | SaWDE | 185.38(−) | 3597 (+) | 95.49 | **94.60±0.7** (−) |
| | | SFE | 110.23(−) | **251** (−) | 94.44 | 92.00±1.1 (+) |
| | | SFE-PSO | 202.8 | 267 | **96.66** | 94.00±0.0 |
| 7 | Carcinom | SM-MEOA | 567.13(+) | 882 (+) | 93.10 | 90.51±0.8 (+) |
| | | SaWDE | 1088.2(+) | 9079 (+) | 95.36 | 93.43±1.2 (+) |
| | | SFE | **147.81**(−) | **223** (−) | 94.84 | 90.86±1.4 (+) |
| | | SFE-PSO | 363.5 | 235 | **99.00** | **96.34±1.6** |
| 8 | Brain Tumor 2 | SM-MEOA | 35.13 (+) | 1163 (+) | 94.46 | 87.00±2.3 (+) |
| | | SaWDE | 155.2 (+) | 10716(+) | 96 | 92.00±2.0 (+) |
| | | SFE | 38.23 (+) | 245 (−) | 98 | 92.00±3.3 (+) |
| | | SFE-PSO | **23.36** | 258 | **100** | **94.46±2.9** |
| 9 | Leukemia 3 | SM-MEOA | **36.03** (−) | 1023 (+) | 97.14 | 93.14±2.1 (+) |
| | | SaWDE | 177 (+) | 10012(+) | **100** | 97.86±1.1 (+) |
| | | SFE | 51.44 (+) | **241** (−) | **100** | 99.00±1.2 (≈) |
| | | SFE-PSO | 119.93 | 255 | **100** | **99.06±1.7** |
| 10 | 11-Tumors | SM-MEOA | 420.43(≈) | 1349 (+) | 86.82 | 85.07±0.8 (+) |
| | | SaWDE | 1114.4(+) | 12965(+) | 92.45 | 90.12±1.4 (≈) |
| | | SFE | **141.03**(−) | **352** (−) | 92.52 | 88.54±2.0 (+) |
| | | SFE-PSO | 416.16 | 273 | **93.09** | **90.29±1.2** |
| 11 | CLL SUB_111 | SM-MEOA | 65.46 (+) | 1144 (+) | 84.66 | 78.04±2.3 (+) |
| | | SaWDE | 1577.2(+) | 11692(+) | 90.00 | 83.02±3.2 (+) |
| | | SFE | 55.56 (+) | 243 (−) | 94.70 | **90.81±3.0** (−) |
| | | SFE-PSO | **48.36** | 251 | **95.57** | 88.59±2.9 |
| Wilcoxon Test(+ \| ≈ \| −) | | SM-MEOA | 3 \| 2 \| 6 | 11 \| 0 \| 0 | − | 8 \| 3 \| 0 |
| | | SaWDE | 9 \| 0 \| 1 | 11 \| 0 \| 0 | − | 4 \| 6 \| 1 |
| | | SFE | 4 \| 0 \| 7 | 0 \| 0 \| 11 | − | 5 \| 4 \| 2 |
| | | SFE-PSO | | | | |
| Friedman Test (Mean Rank) | | SM-MEOA | **1.73** | 3.00 | 3.18 | 3.64 |
| | | SaWDE | 3.91 | 4.00 | 2.64 | 2.36 |
| | | SFE | 1.91 | **1.00** | 2.55 | 2.59 |
| | | SFE-PSO | 2.45 | 2.00 | **1.64** | **1.41** |

As illustrated in Table II, the running time of the SFE-PSO is lower than that of the SM-MOEA and SaWDE on 11 datasets out of 11 datasets. The results of the Wilcoxon and Friedman test have also demonstrated the superiority of the SFE-PSO algorithm since it is always significantly better than SM-MOEA and SaWDE algorithms in all datasets. These results demonstrate the low computational cost of the SFE-PSO algorithm over the SM-MOEA and SaWDE algorithms. It should be noted that the running time of SFE and SFE-PSO algorithms in all datasets is similar. Meanwhile, the running time of the SM-MOEA and SaWDE algorithms increases dramatically with the increase in the number of dataset features. In addition, the running time of the SaWDE algorithm in all datasets, especially on extremely high-dimensional datasets such as 11-Tumors and CLL SUB_111 is significantly high. For example, the running time of the SaWDE algorithm in the CLL SUB_111 dataset is about 47 times the running time of the SFE-PSO algorithm.

The number of features selected by the SFE-PSO is higher than that of the SM-MOEA algorithm. The Wilcoxon and Friedman test results have also demonstrated the superiority of the SM-MOEA algorithm over the SFE-PSO algorithm. The reason for the good performance in terms of the number of selected features by the SM-MOEA algorithm is the use of the dimensionality reduction operator. However, we note that the wrong removal of some relevant features from the dataset can cause the poor performance in terms of the classification accuracy. Therefore, based on statistical tests, the average classification accuracy of the SM-MOEA algorithm is worse than that of the SFE-PSO algorithm in 8 out of 11 datasets. In the other three datasets, the performance of the two algorithms is similar.

Moreover, Table III shows the comparison of the results of the proposed algorithm with that of the HFS-C-P algorithm. As can be seen from Table III, the mean classification accuracy of SFE-PSO using the SVM classifier is higher than that of the HFS-C-P algorithm on 5 out of 8 datasets. The results of the two algorithms are similar in the two datasets, and only in the Leukemia 3 dataset the HFS-C-P algorithm performs better than the SFE-PSO algorithm. The Friedman test results have also demonstrated the superiority of the SFE-PSO algorithm over the HFS-C-P algorithm. On the contrary, the results show the similar performance of both algorithms using the KNN classifier.

TABLE III
COMPARED RESULTS OF SFE, SFE-PSO AND HFS-C-P [31]

| No. | Dataset | Algorithm | KNN | | | SVM | | |
|---|---|---|---|---|---|---|---|---|
| | | | Size | Best | Mean±Std | Size | Best | Mean±Std |
| 1 | Yale | HFS-C-P | – | – | 71.34 | 357.4 | – | 79.52 |
| | | SFE | 39.6 | 76.39 | 69.2±1.7 | 85.21 | 76.28 | 75.1±1.0 |
| | | SFE-PSO | 64.0 | 78.19 | 73.3±1.7 | 76.38 | 82.89 | 80.4±1.9 |
| 2 | warpAR10P | HFS-C-P | – | – | 83.98 | 198.40 | – | 91.1±2.7 |
| | | SFE | 33.4 | 99.23 | 87.6±5.0 | 323.10 | 100 | 100±0.0 |
| | | SFE-PSO | 34.7 | 94.61 | 84.1±6.6 | 470.23 | 100 | 100±0.0 |
| 3 | Colon | HFS-C-P | – | – | 94.77 | 155 | – | 92.47 |
| | | SFE | 17.3 | 96.66 | 93.2±2.4 | 15.93 | 100 | 95.6±2.0 |
| | | SFE-PSO | 25.1 | 96.66 | 93.8±2.0 | 19.70 | 100 | 95.3±3.0 |
| 4 | SRBCT | HFS-C-P | – | – | 100 | 99.4 | – | 100 |
| | | SFE | 37.8 | 100 | 98.8±1.0 | 45.32 | 100 | 100±0.0 |
| | | SFE-PSO | 55.1 | 100 | 99.1±0.8 | 36.19 | 100 | 100±0.0 |
| 5 | DLBCL | HFS-C-P | – | – | 92.61 | 149.86 | – | 89.6±1.8 |
| | | SFE | 28.6 | 100 | 98.4±1.2 | 29.8 | 99.00 | 97.0±1.4 |
| | | SFE-PSO | 38.2 | 100 | 99.1±1.0 | 23.7 | 99.00 | 96.3±1.6 |
| 6 | CNS | HFS-C-P | – | – | 82.54 | 367.6 | – | 85.91 |
| | | SFE | 37.0 | 96.66 | 87.8±4.2 | 36.93 | 98.33 | 93.5±4.2 |
| | | SFE-PSO | 57.8 | 91.66 | 86.7±4.2 | 69.20 | 100 | 95.5±3.7 |
| 7 | Leukemia 3 | HFS-C-P | – | – | 100 | 152.7 | – | 100 |
| | | SFE | 68.8 | 100 | 98.2±1.4 | 49.66 | 100 | 95.1±3.1 |
| | | SFE-PSO | 99.36 | 100 | 98.4±1.3 | 84.13 | 100 | 97.6±2.0 |
| 8 | Ovarian Cancer | HFS-C-P | – | – | 99.68 | 42.50 | – | 100 |
| | | SFE | 38.66 | 100 | 99.1±0.7 | 51.03 | 100 | 100±0.0 |
| | | SFE-PSO | 39.22 | 100 | 99.6±0.8 | 41.02 | 100 | 100±0.0 |
| | Friedman Test (Mean Rank) | HFS-C-P | – | – | 1.88 | 2.63 | – | 2.38 |
| | | SFE | – | – | 2.38 | 1.75 | – | 1.94 |
| | | SFE-PSO | – | – | 1.75 | 1.63 | – | 1.69 |

As shown in Table III, using the SVM classifier, the number of selected features of the SFE-PSO is smaller than that of the HFS-C-P algorithm. The Friedman test results have also demonstrated the superiority of the SFE-PSO algorithm over the HFS-C-P algorithm. It should be noted that the best classification accuracy using SVM and KNN classifiers and the number of selected features using KNN classifiers are not reported in HFS-C-P [31].

## VI. FURTHER ANALYSIS

To further analyze the strengths and weaknesses of the proposed algorithms, we performed various experiments to understand the effect of the parameters, the number of selected features for the non-selection and selection operators in the SFE algorithm, as well as the time of using EC methods in the hybrid SFE-ECs algorithm. In this section, we discuss the impact of the non-selection operator and the appropriate time to use EC methods to continue the search process.

### A. Effect of non-selection operator

Using an appropriate value for the $UN$ parameter of the non-selection operator can significantly increase the efficiency of the SFE algorithm. In general, the value of $UN$ should be large at the beginning of the search process, and it should be small at the end of the search process so that the algorithm enters the exploration phase to search the entire search space in the early stages of the FS process, and enters the exploitation phase in the final stage of the FS process. To achieve this goal, Equation (1) was used in the SFE algorithm to determine the number selected features for the non-selection operator.

The linear decrease of the $UR$ parameter from 0.3 to 0.001 resulted to achieve the desired goal, which is performing exploration in the early stages of the search process and to perform exploitation in the final stages of the search process. However, using this method in all datasets may not provide the appropriate balance between exploration and exploitation capabilities. Besides, in different datasets different levels of exploration and exploitation are needed to search their search space. In addition to the $UR$ value, determining the $UN$ value using other methods can be effective in various datasets. For this purpose, in addition to Equation (1), another method for determining the $UN$ is used as presented in Equation (4):

$$UN = \left\lceil \frac{Rand \times nvar}{k} \right\rceil \qquad (4)$$

where $Rand$ is a random number with a uniform distribution between 0 and 1, $k$ is an integer number that is randomly selected from the range $[1, N]$. In this section, an experiment was performed on five datasets to evaluate Equation (4) for determining $UN$ values. Figure 8 shows the convergence diagram of the SFE algorithm with the method presented in this section to determine the $UN$ value and the method used in previous experiments. In the analysis performed in this section, $N$ is 20. Therefore, whenever a non-selection operator is applied to the $X$, a random number between 1 and 20 is generated for $k$. As illustrated in Figure 8, the equations presented for determining the $UN$ of the non-selection operator have increased the efficiency of the SFE algorithm. Comparing the behavior of the SFE algorithm with the two different $UN$ values obtained from Equation 1 and Equation 4 indicated that the SFE algorithm does not trap in local optimal points, and the algorithm is converged to better solution.

TABLE IV
COMPARED RESULTS OF TWO METHODS FOR DETERMINING $UN$

| No. | Dataset | Algorithm | Size | Worst | Best | Mean±Std |
|---|---|---|---|---|---|---|
| 1 | Yale | SFE (Eq. 1) | 50.27 | 67.87 | 72.12 | 70.41± 1.20 |
| | | SFE (Eq. 4) | 134.90 | 69.09 | 75.75 | 73.31± 1.52 |
| 2 | Lymphoma | SFE (Eq. 1) | 84.63 | 95.84 | 98.94 | 97.28±0.95 |
| | | SFE (Eq. 4) | 206.83 | 97.94 | 100 | 98.67±0.77 |
| 3 | TOX_171 | SFE (Eq. 1) | 106.1 | 94.13 | 99.42 | 97.44±1.50 |
| | | SFE (Eq. 4) | 205.43 | 96.50 | 100 | 98.55±0.95 |
| 4 | childhood ALL | SFE (Eq. 1) | 1223 | 35 | 80 | 50.66±17.1 |
| | | SFE (Eq. 4) | 34.76 | 61.66 | 85 | 73.83±5.70 |
| 5 | Carcinom | SFE (Eq. 1) | 147.81 | 86.16 | 94.84 | 90.86 ±2.25 |
| | | SFE (Eq. 4) | 422.16 | 89.04 | 94.26 | 91.73± 1.34 |

Table IV shows the performance of the SFE algorithm with different evaluation criteria for both methods. The comparison illustrates the effect of the methods presented in this section on increasing the efficiency of the SFE algorithm. For example, in the ALL-childhood dataset, the SFE algorithm improved the average classification accuracy of the algorithm by more than 23%. Therefore, this method can significantly improve the performance of the SFE algorithm in some datasets.

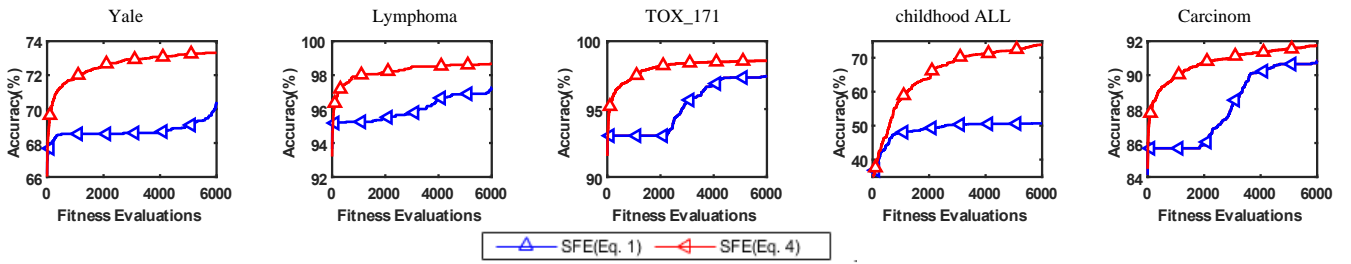

Fig. 8. Effect of two different methods to determine the $UN$ in non-selection operator in the convergence of SFE algorithm.

### B. The effect of time to use EC methods in the SFE-PSO

Determining the appropriate condition for using EC methods to search under the new sub space, and continuing the search process to achieve better solutions can play a critical role in increasing the efficiency of the framework presented in Algorithm 3. If the SFE algorithm does not sufficiently reduce the size of the search space, EC methods may not be able to search the new space due to the large size of the new search space. For example, if a dataset has 20,000 features and the SFE algorithm selects 4,000 features, the search space is still large and the EC methods may not produce desire results. On the other hand, as some important features may be removed from the dataset by the SFE algorithm while searching to reach this reduced search space, the EC methods may not produce expected results in new sub space. For example, analyzing the results of the SFE algorithm on the Colon dataset indicates that the algorithm can achieve more than 90% classification accuracy with less than 15 features. Therefore, if the SFE algorithm does not improve the solution found, the use of EC methods may not be helpful. Therefore, determining a suitable condition for completing the SFE search process and then using EC methods can play an important role in increasing the efficiency of the Framework No. 3.

### C. Computational complexity

The computational complexity of ECs methods such as PSO and Binary Butterfly Optimization (BOA) [45] is of $O\ (it_{max}\ (D \times NP\ +\ C \times NP\ ))$ where $it_{max}$ represents the maximum number of iterations, $D$ is the number of dataset dimensions (i.e., number of features), $C$ is the computational cost of the fitness function, and $NP$ is the population size. However, the computational complexity of other ECs methods such as GA and GWO is of $O\ (it_{max} \times (D \times NP\ +\ C \times NP\ +\ NP\ log\ (NP\ )))$ in the best case and $O\ (it_{max}\ \times (D \times NP\ +\ C \times NP\ +\ NP^2\ )$ in the worst case. This means that the computational complexity of GA and GWO is worse than those of PSO and BOA due to the need to sort the solutions out in each iteration [45].

The computational complexity of the SFE algorithm is of $O\ (it_{max}\ (D\ +\ C\ ))$. This time complexity is better than all ECs methods. In addition, not selecting a large number of features in the early stages of the search process reduces the computational cost of the fitness function, and the computational cost of the algorithm is significantly reduced as well.

The computational complexity of SFE-PSO algorithm is of $O\ (SFE_{it}\ (D\ +\ C\ ) + PSO_{it}(D \times NP\ +\ C \times NP\ ))$, where $SFE_{it}$ is the number of iterations performed in the SFE algorithm, and $PSO_{it}$ is the number of iterations performed by the PSO algorithm. We should also pay attention to the fact that $it_{max} = SFE_{it} + PSO_{it}$. This complexity means that from the maximum number of fitness function evaluations (number of iterations) determined for the search process of the SFE-PSO algorithm, the $SFE_{it}$ iterations are performed by the SFE algorithm, and the $PSO_{it}$ iterations by the PSO algorithm. A notable point regarding the complexity of the SFE-PSO algorithm is that the SFE algorithm unselects irrelevant features from the solution in the first stage of the search process. Therefore, by the continuation of the search process by the PSO algorithm due to the small search space, the computational cost of searching is significantly reduced compared to searching in the original search space. Therefore, the computational cost of the SFE-PSO algorithm is generally significantly lower than all EC algorithms. This issue is evident in Table II and Table VI in the supplementary material, which show the running time in the experiments performed.

## VII. CONCLUSION

In this paper, a FS algorithm called SFE is presented for high-dimensional data. The purpose of this algorithm is to achieve the highest classification accuracy with the least number of features, and the lowest computational cost and memory consumption in high-dimensional data. The experimental results indicate the superiority of the proposed SFE algorithm in compared to other algorithms proposed for the same purpose. Despite the high efficiency of the SFE algorithm in achieving FS goals in high-dimensional data, in some datasets the SFE algorithm may stop at local optimal points after reducing the data dimensionality. To overcome this issue, a hybrid algorithm based on both SFE algorithm and PSO algorithm was proposed. The results of SFE-PSO hybrid algorithm show that this method is more efficient than the SFE algorithm. Therefore, the SFE algorithm can be combined with other EC methods to achieve a higher performance.

Despite the effectiveness and high efficiency of the proposed methods, both of these algorithms have weaknesses that can be fixed in future. One of the weaknesses of the SFE algorithm, which was identified in the behavior analysis of the algorithm (presented in Figure 6) and on some datasets such as ORL, Yala, Carcinom, and TOX_171, is the inappropriateness of the $UN$ and $SN$ parameters in these datasets. To overcome the weakness, there are some mechanisms to design the SFE algorithm with self-organizing parameters. In addition, the non-selection and selection operators proposed in this paper could be improved. Although, the results indicate the high performance of the SFE-PSO algorithm, the wrong removal of relevant features (in the physical removal phase) may cause poor performance and trap the algorithm in local optimal points. In addition, we are able to not remove the features physically, instead, the PSO algorithm is used to perform a local search among the selected features during the different stages of the SFE algorithm search process. Moreover, to achieve high effectiveness and efficiency in the SFE-PSO algorithm, instead of using the PSO algorithm, we can use the improved versions of this algorithm [36] and other EC methods, including the methods proposed in [37]-[42] which significantly improve the efficiency of FS from low-dimensional datasets.

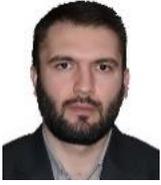 **Behrouz Ahadzadeh** received the M.Sc. degree in Artificial Intelligence and Robotics Engineering from Department of Electrical, Computer and IT Engineering, Qazvin Branch, Islamic Azad University, Qazvin, Iran, in 2014. His research interests include computational intelligence, machine learning, computer vision, knowledge discovery and data mining.

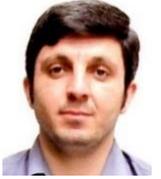 **Moloud Abdar** received the bachelor's degree in computer engineering from Damghan University, Iran, in 2015, and the master's degree in computer science from the University of Aizu, Aizu, Japan, in 2018, and the Ph.D. degree in engineering (computer, machine learning) from the Institute for Intelligent Systems Research and Innovation (IISRI), Deakin University, Australia, in 2022. He is currently an Associate Research Fellow with IISRI at Deakin University, VIC, Australia. His research interests include machine learning, deep learning, computer vision, medical image analysis, sentiment analysis, explainability, and uncertainty quantification.

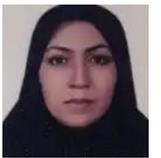 **Fatemeh Safara** received the B.Sc. degree (Hons.) in applied mathematics from Islamic Azad University, and the M.Sc. degree in software engineering from the Tarbiat Modares University, Tehran. She received her Ph.D. degree in Artificial intelligence from University Putra Malaysia (UPM), in 2014. She was a Researcher in the Iran Telecommunication Research Center, Information Technology Department, from 2000 to 2010, doing research on image mining and biometric signals. She joined Islamic Azad University, Islamshahr Branch, on 2010, where she is currently an Assistant Professor with the faculty of computer engineering. Her current research interests include machine learning, data mining, deep learning, signal processing and in particular biological signal processing, medical image analysis, and the Internet of Medical Things.

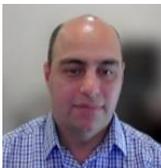 **Abbas Khosravi** (Senior Member, IEEE) received the Ph.D. degree in machine learning from Deakin University, Waurn Ponds, VIC, Australia, in 2010. He is currently an Associate Professor with the Institute for Intelligent Systems Research and Innovation, Deakin University. His broad research interests include artificial intelligence, deep learning, and uncertainty quantification. He is currently researching and applying probabilistic deep learning ideas and uncertainty-aware solutions in healthcare and engineering domains.

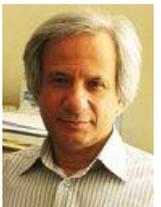 **Mohammad Bagher Menhaj** received the Ph.D. degree from Oklahoma State University (OSU), Stillwater, in 1992. Then, he became a Postdoctoral Fellow with OSU. In 1993, he joined Amirkabir University of Technology, Tehran, Iran. From December 2000 to August 2003, he was with the School of Electrical and Computer Engineering, Department of Computer Science, OSU, as a Visiting Faculty Member and a Research Scholar. He has over 500 publications. His research interests include computational intelligence, learning automata, power systems, image processing, pattern recognition, communications, cognitive science and knowledge discovery.

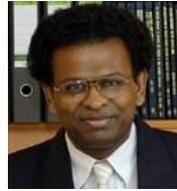 **Ponnuthurai Nagaratnam Suganthan** received the B.A degree and M.A degree in Electrical and Information Engineering from the University of Cambridge, UK in 1990, 1992 and 1994, respectively. He received an honorary doctorate (i.e. Doctor Honoris Causa) in 2020 from University of Maribor, Slovenia. After completing his PhD research in 1995, he served as a pre-doctoral Research Assistant in the Dept of Electrical Engineering, University of Sydney in 1995–96 and a lecturer in the Dept of Computer Science and Electrical Engineering, University of Queensland in 1996–99. Since August 2022, he has been with KINDI Centre for Computing Research, Qatar University, as a research professor. He was an Editorial Board Member of the Evolutionary Computation Journal, MIT Press (2013-2018). He is/was an associate editor of the Applied Soft Computing (Elsevier, 2018-), Neurocomputing (Elsevier, 2018-), IEEE Trans on Cybernetics (2012 - 2018), IEEE Trans on Evolutionary Computation (2005 - 2021), Information Sciences (Elsevier) (2009 - ), Pattern Recognition (Elsevier) (2001 - ) and IEEE Trans on SMC: Systems (2020 - ) Journals. He is a founding co-editor-in-chief of Swarm and Evolutionary Computation (2010 - ), an SCI Indexed Elsevier Journal. His co-authored SaDE paper (published in April 2009) won the "IEEE Trans. on Evolutionary Computation outstanding paper award" in 2012. His research interests include randomization-based learning methods, swarm and evolutionary algorithms, pattern recognition, deep learning and applications of swarm, evolutionary & machine learning algorithms. He was selected as one of the highly cited researchers by Thomson Reuters Science Citations yearly from 2015 to 2022 in computer science. He served as the General Chair of the IEEE SSCI 2013. He has been a member of the IEEE (S'91, M'92, SM'00, Fellow 2015) since 1991 and an elected AdCom member of the IEEE Computational Intelligence Society (CIS) in 2014-2016. He was an IEEE CIS distinguished lecturer (DLP) in 2018-2021.

# Supplementary Material for "SFE: A Simple, Fast and Efficient Feature Selection Algorithm for High-Dimensional Data"

Behrouz Ahadzadeh, Moloud Abdar, Fatemeh Safara, Abbas Khosravi, *Senior Member, IEEE*, Mohammad Bagher Menhaj, Ponnuthurai Nagaratnam Suganthan, *Fellow, IEEE*

## I. EXPERIMENT DESIGN

### A. Datasets

In order to evaluate the effectiveness and efficiency of the proposed FS methods, experiments have been performed on 40 real datasets as presented in Table II. These datasets can be downloaded from https://data.mendeley.com/datasets/fhx5zgx 2zj/1, https://ckzixf.github.io/dataset.html, http://csse.szu.edu. cn/staff/zhuzx/datasets.html, https://jundongl.github.io/scikitfe ature/datasets.html, https://figshare.com/articles/dataset/Micro array_data_rar/7345880/2, https://file.biolab.si/biolab/supp/bic ancer/projections/info/gastricGSE2685.html.

### B. Comparative algorithms and parameter settings

Nine algorithms have been used to compare with the proposed FS methods. First, the results of SFE and SFE-PSO algorithms are compared with the six algorithms BDE [43], BPSO [44], BGA [5], and three state-of-the-art EC-based feature selection algorithms, namely DimReM-BDE [5], DimReM-BGA [5], and DimReM-PSO [5]. Table I lists the details of the parameter settings for these algorithms. The same 40 datasets of Table II are used for all experiments. Then, the results of the proposed algorithms are compared with three state-of-the-art EC-based feature selection algorithms, namely, HFS-C-P [31], SaWDE [32], and SM-MOEA [33] algorithms.

#### TABLE I
#### PARAMETER SETTING

| Algorithms | Parameter values |
|---|---|
| BDE [43] and DimReM-BDE [5] | F = 0.8; CR = 0.2; b = 20 |
| BPSO [44] and DimReM-BPSO [5] | w = 1; c1 = 2; c2 = 1.5 |
| GA [5] and DimReM-GA [5] | pc = 0.6; pm = 0.03 |
| SFE and SFE-PSO | $UR_{max} = 0.3$; $UR_{mIN} = 0.001$, $SN = 1$ |

The reason for choosing the three algorithms BDE [43], BPSO [44], and BGA [5], is to evaluate the performance of these algorithms as three well-known EC algorithms in FS on high-dimensional data. Also, due to the high efficiency of three algorithms, DimReM-BDE [5], DimReM-BGA [5], and DimReM-PSO [5], in reducing the size of high-dimensional data and the availability of their source codes, these algorithms also compared with the proposed method. The three algorithms HFS-C-P [31], SaWDE [32], and SM-MOEA [33] are chosen since they were recently proposed for feature selection in the high-dimensional dataset. In the first part of the experiments, 6000 fitness evaluations were performed. In addition, the population size of each of the six algorithms is 20. In the FS process, k-nearest neighbor (KNN) with K=1 is used as a classifier in evaluating the solutions found by the algorithms. Then 5-fold cross-validation method is used to divide the dataset into train and test sets. Each algorithm has been run 30 times independently. Then, the average classification accuracies, the average number of selected features, and the running time of each algorithm are calculated and compared. All experiments in this paper are carried out on Intel (R) core (TM) i5-5200 U CPU, 2.2 GHz, 12.00 GB RAM, and implemented in MATLAB R2019a. This section evaluates the performance of SFE and SFE-PSO algorithms and compares them with that of BDE, BPSO, BGA, DimReM-BDE, DimReM-BGA, and DimReM-BPSO.

#### TABLE II
#### BASIC INFORMATION OF THE 40 DATASETS

| .No | Dataset | #Ins. | #Fea. | #Classes | Area |
|---|---|---|---|---|---|
| 1 | ORL | 400 | 1024 | 40 | Face Image |
| 2 | Yale | 165 | 1024 | 15 | Face Image |
| 3 | colon | 62 | 2000 | 2 | Biological |
| 4 | SRBCT | 83 | 2308 | 4 | Biological |
| 5 | warpAR10P | 130 | 2400 | 10 | Face Image |
| 6 | Lung Cancer 1 | 203 | 3312 | 5 | Biological |
| 7 | lymphoma | 96 | 4026 | 9 | Biological |
| 8 | GLIOMA | 50 | 4434 | 4 | Biological |
| 9 | Leukemia 1 | 72 | 5327 | 3 | Biological |
| 10 | DLBCL | 77 | 5469 | 2 | Biological |
| 11 | bladder cancer | 40 | 5724 | 3 | Biological |
| 12 | 9-Tumors | 60 | 5726 | 9 | Biological |
| 13 | TOX_171 | 171 | 5748 | 4 | Biological |
| 14 | Brain_Tumor1 | 90 | 5920 | 5 | Biological |
| 15 | Prostate_GE | 102 | 5966 | 2 | Biological |
| 16 | leukemia | 72 | 7070 | 2 | Biological |
| 17 | ALLAML | 72 | 7129 | 2 | Biological |
| 18 | CNS | 60 | 7129 | 2 | Biological |
| 19 | ALL-AML-3 | 72 | 7129 | 3 | Biological |
| 20 | ALL-AML-4 | 72 | 7129 | 4 | Biological |
| 21 | Leukemia 2 | 72 | 7129 | 4 | Biological |
| 22 | childhood ALL | 60 | 8280 | 4 | Biological |
| 23 | Carcinom | 174 | 9182 | 11 | Biological |
| 24 | nci9 | 60 | 9712 | 9 | Biological |
| 25 | arcene | 200 | 10000 | 2 | Biological |
| 26 | orlraws10P | 100 | 10304 | 10 | Face Image |
| 27 | Brain_Tumor2 | 50 | 10367 | 4 | Biological |
| 28 | Leukemia 3 | 72 | 11225 | 3 | Biological |
| 29 | CLL SUB_111 | 111 | 11340 | 3 | Biological |
| 30 | 11-Tumors | 174 | 12533 | 11 | Biological |
| 31 | MLL | 72 | 12582 | 3 | Biological |
| 32 | Lung Cancer 2 | 203 | 12600 | 5 | Biological |
| 33 | CML treatment | 28 | 12625 | 2 | Biological |
| 34 | glioblastoma | 50 | 12625 | 4 | Biological |
| 35 | AML prognosis | 54 | 12625 | 2 | Biological |
| 36 | prostate cancer | 20 | 12627 | 2 | Biological |
| 37 | Ovarian Cancer | 253 | 15154 | 2 | Biological |
| 38 | SMK_CAN_187 | 187 | 19993 | 2 | Biological |
| 39 | GLI_85 | 85 | 22283 | 2 | Biological |
| 40 | Breast Cancer | 97 | 24481 | 2 | Biological |

TABLE III
THE WORST, BEST, MEAN AND STD OF THE CLASSIFICATION ACCURACY OBTAINED BY EIGHT ALGORITHMS OVER THE 30
INDEPENDENT RUNS

| No. | Dataset | Metrics | BDE [43] | GA [5] | BPSO [44] | DimReM-BDE [5] | DimReM-BGA [5] | DimReM-BPSO [5] | SFE | SFE-PSO |
|---|---|---|---|---|---|---|---|---|---|---|
| 1 | ORL | Worst | 96.2500 | 96.0000 | 97.0000 | 96.5000 | 96.2500 | 96.5000 | 94.5000 | 96.2500 |
| | | Best | 96.7500 | 97.0000 | 97.7500 | 97.2500 | 97.0000 | 97.5000 | 97.0000 | 98.2500 |
| | | Mean | 96.4900(+) | 96.5500(+) | **97.3500**(≈) | 96.8600(+) | 96.7396(+) | 97.0500(≈) | 95.7167(+) | 97.2417 |
| | | Std | 0.18370 | 0.27000 | 0.2282 | 0.1920 | 0.2270 | 0.3146 | 0.6612 | 0.5704 |
| 2 | Yale | Worst | 73.3333 | 72.1212 | 74.5455 | 73.9394 | 73.9394 | 73.9394 | 67.8788 | 71.5152 |
| | | Best | 75.1515 | 75.1515 | 75.7576 | 75.1515 | 75.7576 | 75.7576 | 72.1212 | 77.5758 |
| | | Mean | 73.8292(+) | 74.0220(+) | 75.1515(≈) | 74.8209(≈) | 74.5730(≈) | 74.9587(≈) | 70.4132(+) | **75.2323** |
| | | Std | 0.40270 | 0.73150 | 0.2645 | 0.3611 | 0.5114 | 0.3917 | 1.2069 | 1.3574 |
| 3 | Colon | Worst | 77.3077 | 78.8462 | 80.5128 | 78.8462 | 78.8462 | 78.8462 | 90.5128 | 91.9231 |
| | | Best | 82.1795 | 82.1795 | 85.5128 | 82.1795 | 85.5128 | 85.5128 | 100.000 | 100.000 |
| | | Mean | 79.3034(+) | 80.4615(+) | 82.8462(+) | 79.9060(+) | 81.2393(+) | 81.3462(+) | 96.4402(≈) | **96.5128** |
| | | Std | 0.91200 | 1.19160 | 1.2066 | 0.9913 | 1.5904 | 1.5538 | 2.4443 | 2.4659 |
| 4 | SRBCT | Worst | 90.5147 | 90.5147 | 94.0441 | 91.6912 | 90.5147 | 91.6912 | 92.7941 | 96.4706 |
| | | Best | 94.1176 | 96.4706 | 96.0319 | 94.1176 | 96.4706 | 96.4706 | 100.000 | 100.000 |
| | | Mean | 91.9461(+) | 93.0490(+) | 91.6471(+) | 92.6618(+) | 93.8480(+) | 94.2721(+) | 98.8971(≈) | **99.0931** |
| | | Std | 0.7286 | 1.42760 | 0.7979 | 0.6463 | 1.6069 | 1.2075 | 1.9921 | 1.1494 |
| 5 | warpAR10P | Worst | 59.2308 | 59.2308 | 63.0769 | 60.7692 | 60.0000 | 61.5385 | 83.0729 | 92.3077 |
| | | Best | 62.3077 | 61.4359 | 65.3077 | 63.0769 | 62.0769 | 66.9231 | 99.2308 | 99.2308 |
| | | Mean | 60.5641(+) | 63.8462(+) | 68.4618(+) | 61.8974(+) | 63.3846(+) | 63.8718(+) | 93.6410(+) | **95.5385** |
| | | Std | 0.72650 | 1.10150 | 1.2510 | 0.6920 | 1.1805 | 1.2853 | 3.3919 | 2.2979 |
| 6 | Lung Cancer 1 | Worst | 96.5732 | 96.0854 | 96.5732 | 96.5732 | 96.0854 | 96.0854 | 96.0854 | 97.5610 |
| | | Best | 97.0732 | 97.5610 | 97.5610 | 97.0732 | 97.5610 | 97.5610 | 98.5366 | 99.0244 |
| | | Mean | 96.7069(+) | 96.7374(+) | 97.2472(+) | 96.8199(+) | 96.8049(+) | 96.9358(+) | 97.4122(≈) | **98.3252** |
| | | Std | 0.2219 | 0.29770 | 0.3309 | 0.2485 | 0.3826 | 0.3649 | 0.6435 | 0.3775 |
| 7 | lymphoma | Worst | 96.8947 | 96.8947 | 96.8947 | 96.8947 | 96.8947 | 96.8947 | 95.8421 | 96.8947 |
| | | Best | 97.9474 | 97.9474 | 99.0000 | 97.9474 | 97.9474 | 97.9474 | 98.9474 | 100.000 |
| | | Mean | 97.0088(+) | 97.0351(+) | 97.4211(+) | 97.1175(+) | 97.1421(+) | 97.2825(+) | 97.2860(+) | **99.3526** |
| | | Std | 0.31880 | 0.36390 | 0.6025 | 0.4226 | 0.4519 | 0.5147 | 0.9554 | 0.9532 |
| 8 | GLIOMA | Worst | 84.0000 | 84.0000 | 84.0000 | 84.0000 | 84.0000 | 84.0000 | 92.0000 | 92.0000 |
| | | Best | 88.0000 | 88.0000 | 92.0000 | 90.0000 | 90.0000 | 90.0000 | 100.000 | 100.000 |
| | | Mean | 84.5333(+) | 87.7333(+) | 89.1333(+) | 84.6667(+) | 87.6667(+) | 87.5333(+) | **97.0000**(≈) | 96.1333 |
| | | Std | 1.04170 | 1.55220 | 1.7167 | 0.9589 | 1.6678 | 2.0126 | 2.6130 | 2.0965 |
| 9 | Leukemia 1 | Worst | 93.1429 | 93.0476 | 95.9048 | 93.1429 | 93.0476 | 94.4762 | 94.5714 | 97.2381 |
| | | Best | 95.9048 | 95.9048 | 97.3333 | 95.9048 | 97.3333 | 97.3333 | 100.000 | 100.000 |
| | | Mean | 94.0667(+) | 94.6000(+) | 96.5746(+) | 94.2508(+) | 94.7429(+) | 95.4952(+) | 98.2825(≈) | **98.8794** |
| | | Std | 0.82960 | 1.08990 | 0.6815 | 0.8394 | 0.9781 | 0.9748 | 1.3848 | 0.7234 |
| 10 | DLBCL | Worst | 89.7500 | 88.5000 | 89.7500 | 88.5000 | 89.7500 | 88.5000 | 95.0000 | 96.2500 |
| | | Best | 89.8333 | 92.3333 | 92.3333 | 91.0000 | 89.9194 | 90.5083 | 100.000 | 100.000 |
| | | Mean | 89.7639(+) | 90.4667(+) | 91.0500(+) | 89.9194(+) | 90.5083(+) | 90.1722(+) | 99.1611(≈) | **99.3639** |
| | | Std | 0.0316 | 0.8629 | 0.7733 | 0.6337 | 0.7828 | 1.0079 | 1.1075 | 1.0393 |
| 11 | bladder cancer | Worst | 75.0000 | 75.0000 | 75.0000 | 75.0000 | 75.0000 | 75.0000 | 92.5000 | 90.0000 |
| | | Best | 82.5000 | 90.0000 | 90.0000 | 82.5000 | 90.0000 | 90.0000 | 100.000 | 100.000 |
| | | Mean | 78.4000(+) | 84.0000(+) | 85.9000(+) | 78.4000(+) | 84.0000(+) | 82.8000(+) | **97.5000**(≈) | 97.3000 |
| | | Std | 2.4875 | 3.9528 | 4.8348 | 2.0259 | 4.6256 | 5.4639 | 2.1651 | 2.6926 |
| 12 | 9-Tumors | Worst | 51.6667 | 55.0000 | 56.6667 | 53.3333 | 53.3333 | 53.3333 | 56.6667 | 65.0000 |
| | | Best | 56.6667 | 58.3333 | 61.6667 | 56.6667 | 58.3333 | 61.6667 | 86.6667 | 85.0000 |
| | | Mean | 54.2222(+) | 56.2222(+) | 59.2222(+) | 54.1667(+) | 56.2778(+) | 56.8889(+) | 72.7778(+) | **74.1667** |
| | | Std | 1.3655 | 1.3795 | 1.2935 | 0.9539 | 1.4306 | 1.7904 | 7.8092 | 4.9276 |
| 13 | TOX_171 | Worst | 95.8992 | 95.3109 | 97.6471 | 95.8992 | 95.8992 | 96.4874 | 94.1345 | 98.8235 |
| | | Best | 97.0588 | 97.6639 | 98.8235 | 97.6471 | 98.8235 | 98.4286 | 99.4286 | 100.000 |
| | | Mean | 96.4443(+) | 96.7356(+) | 98.5490 | 96.5445(+) | 96.8717(+) | 97.3776(+) | 97.4420(+) | **99.7849** |
| | | Std | 0.40050 | 0.5460 | 0.4296 | 0.4669 | 0.6196 | 0.5248 | 1.5030 | 0.3264 |
| 14 | Brain_Tumor1 | Worst | 90.0000 | 90.0000 | 91.1111 | 90.0000 | 90.0000 | 90.0000 | 90.0000 | 92.2222 |
| | | Best | 91.1111 | 91.1111 | 91.1111 | 91.1111 | 91.1111 | 91.1111 | 94.4444 | 96.6667 |
| | | Mean | 90.3704(+) | 90.9630(+) | 91.1111(+) | 90.3704(+) | 90.8519(+) | 90.9259(+) | 92.0000(≈) | **94.0000** |
| | | Std | 0.53270 | 0.38420 | 2.89e-14 | 0.5327 | 0.4780 | 0.4212 | 1.1451 | 0.9500 |
| 15 | Prostate GE | Worst | 88.1905 | 88.1905 | 89.1429 | 88.1905 | 88.1429 | 88.1905 | 93.1429 | 92.0952 |
| | | Best | 90.0952 | 90.1429 | 92.0952 | 90.1429 | 90.1429 | 90.0952 | 99.0000 | 99.0000 |
| | | Mean | 88.9016(+) | 89.0905(+) | 90.3079(+) | 88.8667(+) | 88.9270(+) | 89.5159(+) | **97.0524**(≈) | 96.3429 |
| | | Std | 0.5000 | 0.6184 | 0.6384 | 0.5184 | 0.6636 | 0.7897 | 1.4538 | 1.6871 |
| 16 | leukemia | Worst | 81.7143 | 81.7143 | 84.5714 | 81.8095 | 81.7143 | 81.7143 | 90.2857 | 97.2381 |
| | | Best | 84.5714 | 86.0000 | 86.1429 | 84.5714 | 87.4286 | 86.0952 | 100.000 | 100.000 |
| | | Mean | 82.9270(+) | 84.0063(+) | 87.4286(+) | 83.2159(+) | 84.2889(+) | 84.9111(+) | 95.8254(+) | **99.0571** |
| | | Std | 0.9066 | 1.1578 | 0.6867 | 0.6793 | 1.1583 | 1.1698 | 2.0384 | 0.7255 |
| 17 | ALLAML | Worst | 90.1905 | 90.1905 | 91.6190 | 90.1905 | 88.8571 | 90.1905 | 95.7143 | 95.9048 |
| | | Best | 91.6190 | 91.6190 | 91.6190 | 91.6190 | 91.6190 | 91.6190 | 100.000 | 100.000 |
| | | Mean | 90.9016(+) | 91.4286(+) | 91.6190(+) | 91.1683(+) | 91.3397(+) | 91.3365(+) | 98.9302(≈) | **99.2222** |
| | | Std | 0.6830 | 0.4939 | 4.33e-14 | 0.6488 | 0.6743 | 0.5750 | 1.1447 | 1.1183 |
| 18 | CNS | Worst | 70.0000 | 71.6667 | 73.3333 | 71.6667 | 71.6667 | 71.6667 | 85.0000 | 85.0000 |
| | | Best | 75.0000 | 75.0000 | 80.0000 | 75.0000 | 78.3333 | 78.3333 | 98.3333 | 96.6667 |
| | | Mean | 73.0000(+) | 72.2222(+) | 76.7778(+) | 73.2778(+) | 74.2222(+) | 74.6111(+) | 91.5000(≈) | **92.6111** |
| | | Std | 1.1073 | 1.0480 | 2.0846 | 0.9268 | 1.2172 | 1.6772 | 3.3434 | 2.7917 |
| 19 | ALL-AML-3 | Worst | 87.3333 | 87.3333 | 88.7619 | 87.3333 | 87.3333 | 87.3333 | 93.0476 | 94.4762 |
| | | Best | 88.7619 | 88.7619 | 90.1905 | 88.7619 | 88.7619 | 90.1905 | 100 | 100 |
| | | Mean | 87.7619(+) | 88.3397(+) | 89.0000(+) | 88.1778(+) | 88.3397(+) | 88.5778(+) | 97.1397(≈) | **97.6857** |
| | | Std | 0.6145 | 0.6564 | 0.5415 | 0.6798 | 0.6564 | 0.6075 | 2.1855 | 1.5575 |
| 20 | ALL-AML-4 | Worst | 84.6667 | 86.0000 | 86.0000 | 84.6667 | 86.0000 | 86.0000 | 88.7619 | 92.8571 |
| | | Best | 86.0000 | 87.4286 | 87.4286 | 86.0000 | 87.4286 | 87.4286 | 98.6667 | 100.000 |
| | | Mean | 85.6889(+) | 86.0952(+) | 86.0952(+) | 85.6889(+) | 86.0476(+) | 86.0952(+) | 94.4063(+) | **96.0095** |
| | | Std | 0.5736 | 0.3624 | 0.3624 | 0.5736 | 0.2608 | 0.3624 | 3.1007 | 1.7693 |

TABLE IV
THE WORST, BEST, MEAN AND STD OF THE CLASSIFICATION ACCURACY OBTAINED BY EIGHT ALGORITHMS OVER THE 30
INDEPENDENT RUNS

| No. | Dataset | Metrics | BDE [43] | BGA [5] | BPSO [44] | DimReM-BDE [5] | DimReM-BGA [5] | DimReM-BPSO [5] | SFE | SFE-PSO |
|---|---|---|---|---|---|---|---|---|---|---|
| 21 | Leukemia 2 | Worst | 84.6667 | 84.6667 | 86.0000 | 84.6667 | 86.0000 | 86.0000 | 88.8571 | 91.6190 |
| | | Best | 86.0952 | 86.0000 | 87.4286 | 87.4286 | 86.0000 | 87.4286 | 98.5714 | 100.000 |
| | | Mean | 85.8286(+) | 85.9556(+) | 86.3810(+) | 85.8698(+) | 86.0000(+) | 86.0179(+) | 94.9794(≈) | **95.6063** |
| | | Std | 0.4641 | 0.2434 | 0.6425 | 0.5459 | 0.000000 | 0.2608 | 2.3460 | 1.7084 |
| 22 | childhood ALL | Worst | 35.0000 | 35.0000 | 35.0000 | 35.0000 | 35.0000 | 35.0000 | 35.0000 | 35.0000 |
| | | Best | 35.0000 | 35.0000 | 35.0000 | 35.0000 | 35.0000 | 35.0000 | 80.0000 | 75.0000 |
| | | Mean | 35.0000(+) | 35.0000(+) | 35.0000(+) | 35.0000(+) | 35.0000(+) | 35.0000(+) | 50.6667(≈) | **51.8000** |
| | | Std | 0.00000 | 0.00000 | 0.00000 | 0.00000 | 0.00000 | 0.00000 | 17.1931 | 15.7301 |
| 23 | Carcinom | Worst | 89.0420 | 89.0420 | 90.7563 | 89.0420 | 89.0420 | 89.6134 | 86.1681 | 94.6179 |
| | | Best | 90.7563 | 91.3445 | 92.4874 | 90.7731 | 91.3277 | 91.9160 | 94.8403 | 100.0000 |
| | | Mean | 90.0817(+) | 90.1100(+) | 91.3636(+) | 90.0863(+) | 90.0817(+) | 90.5019(+) | 90.8625(+) | **96.4706** |
| | | Std | 0.3366 | 0.5962 | 0.4524 | 0.4234 | 0.6017 | 0.5825 | 2.2512 | 0.9897 |
| 24 | nci9 | Worst | 56.6667 | 58.3333 | 60.0000 | 56.6667 | 58.3333 | 58.3333 | 63.3333 | 68.3333 |
| | | Best | 60.0000 | 61.6667 | 65.0000 | 60.0000 | 61.6667 | 61.6667 | 81.6667 | 85.0000 |
| | | Mean | 58.7222(+) | 61.6667(+) | 62.3889(+) | 59.1111(+) | 60.1111(+) | 60.6111(+) | 73.1111(+) | **75.5556** |
| | | Std | 0.9472 | 1.1029 | 1.1315 | 0.9522 | 1.2328 | 1.1145 | 4.6265 | 3.3141 |
| 25 | arcene | Worst | 88.0000 | 88.5000 | 90.0000 | 89.0000 | 88.5000 | 90.0000 | 86.0000 | 90.0000 |
| | | Best | 94.5000 | 93.0000 | 93.0000 | 93.0000 | 92.5000 | 93.0000 | 97.5000 | 97.0000 |
| | | Mean | 89.3667(+) | 90.6667(+) | 91.6957(+) | 89.9000(+) | 90.7167(+) | 91.4000(+) | 92.0167(+) | **93.3667** |
| | | Std | 0.6940 | 1.2058 | 0.6698 | 0.7812 | 0.8272 | 0.6747 | 3.2417 | 1.7564 |
| 26 | orlraws10P | Worst | 98.0000 | 98.0000 | 98.0000 | 98.0000 | 98.0000 | 97.0000 | 97.0000 | 99.0000 |
| | | Best | 98.0000 | 98.0000 | 98.0000 | 99.0000 | 99.0000 | 99.0000 | 100.000 | 100.000 |
| | | Mean | 98.0000(+) | 98.0000(+) | 98.0000(+) | 98.0333(+) | 98.6667(+) | 98.0000(+) | 99.4000(≈) | **99.7300** |
| | | Std | 0.00000 | 0.00000 | 0.00000 | 0.1826 | 0.2537 | 0.3718 | 0.7701 | 0.4498 |
| 27 | Brain_Tumor2 | Worst | 70.0000 | 72.0000 | 74.0000 | 70.0000 | 72.0000 | 70.0000 | 86.0000 | 88.0000 |
| | | Best | 76.0000 | 78.0000 | 78.0000 | 74.0000 | 76.0000 | 78.0000 | 98.0000 | 100.000 |
| | | Mean | 72.4667(+) | 75.0000(+) | 76.0000(+) | 72.7333(+) | 74.6667(+) | 75.2000(+) | 92.0000(+) | **94.4667** |
| | | Std | 1.4559 | 1.1447 | 0.9097 | 1.1121 | 1.0933 | 1.6274 | 3.3631 | 2.9094 |
| 28 | Leukemia 3 | Worst | 88.6667 | 88.6667 | 88.6667 | 88.6667 | 88.6667 | 87.3333 | 95.8095 | 91.5238 |
| | | Best | 90.0952 | 91.5238 | 91.5238 | 90.0952 | 90.0952 | 91.5238 | 100.000 | 100.000 |
| | | Mean | 88.8229(+) | 89.4667(+) | 89.5221(+) | 88.8724(+) | 89.5238(+) | 89.4171(+) | 99.0095(≈) | **99.0635** |
| | | Std | 0.3858 | 0.9295 | 0.9221 | 0.4631 | 0.7143 | 1.0091 | 1.2965 | 1.7068 |
| 29 | CLL_SUB_111 | Worst | 72.0158 | 71.9367 | 74.6245 | 72.0158 | 71.1067 | 72.9249 | 82.8063 | 83.7945 |
| | | Best | 74.6640 | 76.5217 | 78.3399 | 74.6640 | 74.7905 | 77.4308 | 94.7036 | 95.5731 |
| | | Mean | 73.1963(+) | 74.4730(+) | 76.7905(+) | 73.4862(+) | 77.3913(+) | 76.0422(+) | **90.8116**(−) | 88.5942 |
| | | Std | 0.7801 | 1.2780 | 1.0127 | 0.7793 | 1.4713 | 1.0635 | 3.0744 | 2.9313 |
| 30 | 11-Tumors | Worst | 81.5798 | 81.5462 | 85.0084 | 82.1513 | 82.7059 | 82.7227 | 85.0434 | 88.4706 |
| | | Best | 83.8655 | 85.0252 | 86.7395 | 84.4370 | 85.0084 | 86.1681 | 92.5210 | 93.0924 |
| | | Mean | 83.0627(+) | 83.5950(+) | 85.6749(+) | 83.3115(+) | 83.7703(+) | 84.8586(+) | 88.5417(+) | **90.2919** |
| | | Std | 0.5734 | 0.9367 | 0.5281 | 0.6265 | 0.7512 | 0.7343 | 2.0926 | 1.2420 |
| 31 | MLL | Worst | 90.1905 | 90.1905 | 88.8571 | 88.8571 | 90.1905 | 90.1905 | 94.2857 | 95.9048 |
| | | Best | 91.5238 | 92.9524 | 92.9524 | 91.5238 | 92.9524 | 91.6190 | 100.000 | 100.000 |
| | | Mean | 90.4127(+) | 90.8159(+) | 91.2159(+) | 90.4127(+) | 91.0413(+) | 90.9079(+) | 98.5587(≈) | **99.1714** |
| | | Std | 0.5054 | 0.7208 | 0.7651 | 0.6148 | 0.7520 | 0.6830 | 1.5405 | 1.2771 |
| 32 | Lung Cancer 2 | Worst | 93.1463 | 92.6463 | 93.6463 | 92.6585 | 92.6463 | 93.1463 | 95.5976 | 96.0976 |
| | | Best | 94.1463 | 94.1341 | 94.6341 | 93.6463 | 94.1463 | 94.6341 | 98.5366 | 99.0244 |
| | | Mean | 93.4873(+) | 93.3644(+) | 94.0951(+) | 93.3283(+) | 93.4049(+) | 93.6454(+) | **97.3839**(≈) | 97.3797 |
| | | Std | 0.277 | 0.3820 | 0.2812 | 0.2822 | 0.3255 | 0.4070 | 0.7177 | 0.7622 |
| 33 | CML treatment | Worst | 56.6667 | 56.6667 | 56.6667 | 56.6667 | 56.6667 | 56.6667 | 78.0000 | 88.6667 |
| | | Best | 56.6667 | 60.6667 | 60.6667 | 60.6667 | 60.6667 | 60.6667 | 100.000 | 100.000 |
| | | Mean | 56.6667(+) | 57.4667(+) | 57.4667(+) | 56.9600(+) | 57.4400(+) | 57.4667(+) | 89.0133(+) | **95.1733** |
| | | Std | 1.45e-14 | 1.0198 | 1.6330 | 1.0198 | 1.4967 | 1.6330 | 5.9446 | 3.8791 |
| 34 | glioblastoma | Worst | 66.0000 | 68.0000 | 70.0000 | 66.0000 | 68.0000 | 68.0000 | 85.0000 | 86.0000 |
| | | Best | 72.0000 | 72.0000 | 72.0000 | 70.0000 | 72.0000 | 72.0000 | 98.3333 | 98.0000 |
| | | Mean | 78.0800(+) | 70.3200(+) | 71.6800(+) | 67.8400(+) | 70.6400(+) | 70.6400(+) | 91.5000(+) | **92.9600** |
| | | Std | 1.3515 | 1.3760 | 0.7483 | 0.9866 | 1.2543 | 1.3808 | 3.3434 | 3.0616 |
| 35 | AML prognosis | Worst | 66.5455 | 68.3636 | 68.3636 | 66.5455 | 66.5455 | 68.3636 | 82.0000 | 84.9091 |
| | | Best | 70.1818 | 70.1818 | 72.0000 | 70.1818 | 72.0000 | 72.0000 | 94.0000 | 98.1818 |
| | | Mean | 68.2909(+) | 69.0909(+) | 70.2545(+) | 68.3636(+) | 69.3018(+) | 69.6000(+) | 88.3200(+) | **92.5818** |
| | | Std | 0.8266 | 0.9091 | 0.9452 | 1.0099 | 1.3045 | 1.0123 | 3.8592 | 4.0588 |
| 36 | prostate cancer | Worst | 70.0000 | 70.0000 | 65.0000 | 70.0000 | 70.0000 | 65.0000 | 80.0000 | 95.0000 |
| | | Best | 70.0000 | 75.0000 | 75.0000 | 75.0000 | 75.0000 | 75.0000 | 100.000 | 100.000 |
| | | Mean | 70.0000(+) | 71.8000(+) | 71.4000(+) | 70.2000(+) | 71.6000(+) | 70.2000(+) | 95.2000(+) | **99.4000** |
| | | Std | 0.0000 | 2.4495 | 2.7080 | 1.0000 | 2.3805 | 2.2730 | 4.8905 | 1.6583 |
| 37 | Ovarian Cancer | Worst | 95.2784 | 95.2784 | 95.2784 | 95.2784 | 95.2784 | 95.2784 | 97.6157 | 98.0314 |
| | | Best | 95.2784 | 95.2784 | 95.2784 | 95.2784 | 95.2784 | 95.2784 | 100.000 | 100.000 |
| | | Mean | 95.2784(+) | 95.2784(+) | 95.2784(+) | 95.2784(+) | 95.2784(+) | 95.2784(+) | 99.1072(≈) | **99.7642** |
| | | Std | 1.46e-14 | 1.46e-14 | 1.46e-14 | 1.46e-14 | 1.46e-14 | 1.46e-14 | 0.7018 | 0.4208 |
| 38 | SMK_CAN_187 | Worst | 68.9758 | 69.2532 | 70.0427 | 69.5021 | 69.5021 | 70.0427 | 79.1750 | 76.9986 |
| | | Best | 71.1095 | 71.1238 | 71.8625 | 70.5832 | 71.1238 | 72.1764 | 87.7098 | 89.3314 |
| | | Mean | 69.8846(+) | 70.1952(+) | 70.6952(+) | 70.0825(+) | 70.5405(+) | 71.1651(+) | 83.0337(≈) | **83.1318** |
| | | Std | 0.4431 | 0.5214 | 0.4495 | 0.4395 | 0.5703 | 0.5333 | 2.3694 | 3.1881 |
| 39 | GLI_85 | Worst | 90.5882 | 90.5882 | 90.5882 | 90.5882 | 90.5882 | 90.5882 | 94.1176 | 92.3077 |
| | | Best | 91.7647 | 92.9412 | 92.9412 | 91.7647 | 92.9412 | 92.9412 | 98.8235 | 99.2308 |
| | | Mean | 91.5294(+) | 91.8588(+) | 91.6706(+) | 91.4510(+) | 92.2745(+) | 91.4824(+) | 96.5490(+) | **98.1569** |
| | | Std | 0.4786 | 0.7533 | 0.6724 | 0.5291 | 0.6686 | 0.7026 | 1.2714 | 1.2235 |
| 40 | Breast Cancer | Worst | 65.9474 | 66.2514 | 69.1053 | 65.9474 | 68.0526 | 67.0000 | 79.3684 | 80.5263 |
| | | Best | 69.1053 | 69.8564 | 71.2105 | 69.1053 | 70.1579 | 72.2105 | 93.8947 | 94.9474 |
| | | Mean | 66.8658(+) | 68.4532(+) | 70.3053(+) | 66.8658(+) | 68.7816(+) | 69.2132(+) | 85.7018(+) | **86.8351** |
| | | Std | 0.8434 | 0.4565 | 0.7757 | 0.8434 | 0.7531 | 1.2170 | 4.1574 | 3.9705 |

| No. | Dataset | BDE [43] | BGA [5] | BPSO [44] | DimReM-BDE [5] | DimReM-BGA [5] | DimReM-BPSO [5] | SFE | SFE-PSO |
|---|---|---|---|---|---|---|---|---|---|
| 1 | ORL | 497.88(+) | 487.52(+) | 471.68(+) | 427.16(+) | 528.37(+) | 394.64(+) | 94.5 (+) | **57.9** |
| 2 | Yale | 500.13(+) | 491.09(+) | 474.18(+) | 441.81(+) | 424.22(+) | 411 (+) | **50.27**(−) | 57.0667 |
| 3 | colon | 988.13(+) | 945.66(+) | 934.53(+) | 936.5 (+) | 862.93(+) | 865.06(+) | 17.4 (+) | **13.0333** |
| 4 | SRBCT | 1138 (+) | 1109 (+) | 1084 (+) | 1083 (+) | 1032 (+) | 1020 (+) | **32.23**(−) | 43.1 |
| 5 | warpAR10P | 1180 (+) | 1154 (+) | 1146 (+) | 1122 (+) | 1080 (+) | 1077 (+) | 33.23(+) | **26.8667** |
| 6 | Lung Cancer 1 | 1625 (+) | 1593 (+) | 1568 (+) | 1581 (+) | 1510 (+) | 1502 (+) | **77.9** (+) | 127.4 |
| 7 | Lymphoma | 1974 (+) | 1849 (+) | 1909 (+) | 1927 (+) | 1846 (+) | 1854 (+) | **84.63**(−) | 149.9 |
| 8 | GLIOMA | 2188 (+) | 2137 (+) | 2116 (+) | 2139 (+) | 2050 (+) | 2056 (+) | 26.8 (−) | 28.6333 |
| 9 | Leukemia 1 | 2647 (+) | 2593 (+) | 2560 (+) | 2587 (+) | 2531 (+) | 2507 (+) | **46.43**(−) | 115.5 |
| 10 | DLBCL | 2695 (+) | 2650 (+) | 2625 (+) | 2653 (+) | 2561 (+) | 2565 (+) | 29 (+) | **18.9333** |
| 11 | bladder cancer | 2820 (+) | 2767 (+) | 2734 (+) | 2487 (+) | 2669 (+) | 2695 (+) | 31.80(+) | **18.6** |
| 12 | 9-Tumors | 2830 (+) | 2785 (+) | 2766 (+) | 2791 (+) | 2716 (+) | 2715 (+) | **65.83**(−) | 153.5333 |
| 13 | TOX_171 | 2846 (+) | 2806 (+) | 2762 (+) | 2800 (+) | 2746 (+) | 2727 (+) | **106.1**(−) | 221.1333 |
| 14 | Brain_Tumor1 | 2933 (+) | 2864 (+) | 2837 (+) | 2862 (+) | 2788 (+) | 2781 (+) | **110.2**(−) | 202.8 |
| 15 | Prostate_GE | 2958 (+) | 2912 (+) | 2897 (+) | 2908 (+) | 2863 (+) | 2835 (+) | 29.8 (−) | **23.7** |
| 16 | leukemia | 3542 (+) | 3487 (+) | 3458 (+) | 3499 (+) | 3416 (+) | 3422 (+) | **37.26**(−) | 70.8667 |
| 17 | ALLAML | 3541 (+) | 3472 (+) | 3425 (+) | 3493 (+) | 3384 (+) | 3387 (+) | 32.66(+) | **27.8333** |
| 18 | CNS | 3531 (+) | 3484 (+) | 3467 (+) | 3486 (+) | 3403 (+) | 3403 (+) | 36.4 (+) | **30.4667** |
| 19 | ALL-AML-3 | 3536 (+) | 3473 (+) | 3434 (+) | 3491 (+) | 3396 (+) | 3396 (+) | **53.96**(−) | 60.6667 |
| 20 | ALL-AML-4 | 3543 (+) | 3427 (+) | 3465 (+) | 3492 (+) | 3396 (+) | 3382 (+) | **62** (−) | 120.8 |
| 21 | Leukemia 2 | 3537 (+) | 3459 (+) | 3431 (+) | 3487 (+) | 3401 (+) | 3394 (+) | **59.43**(−) | 100.0333 |
| 22 | childhood ALL | 12232 (+) | 12232 (+) | 12232 (+) | 12232 (+) | 12232 (+) | 12232 (+) | 1223 (+) | **1204** |
| 23 | Carcinom | 4563 (+) | 4523 (+) | 4490 (+) | 4523 (+) | 4470 (+) | 4490 (+) | **147.8**(−) | 363.5667 |
| 24 | nci9 | 4815 (+) | 4759 (+) | 4730 (+) | 4785 (+) | 4694 (+) | 4675 (+) | 60.2 (+) | **57.5667** |
| 25 | arcene | 4992 (+) | 4970 (+) | 4978 (+) | 4958 (+) | 4933 (+) | 4926 (+) | **208.6**(−) | 276.5333 |
| 26 | orlraws10P | 5081 (+) | 5014 (+) | 4986 (+) | 5046 (+) | 4941 (+) | 4962 (+) | 33.26(+) | **28.8** |
| 27 | Brain_Tumor2 | 5148 (+) | 5070 (+) | 5036 (+) | 5109 (+) | 4993 (+) | 4995 (+) | 38.23(+) | **23.3667** |
| 28 | Leukemia 3 | 5573 (+) | 5455 (+) | 5440 (+) | 5536 (+) | 5398 (+) | 5400 (+) | **51.44**(−) | 119.9333 |
| 29 | CLL SUB_111 | 5657 (+) | 5629 (+) | 5646 (+) | 5614 (+) | 5600 (+) | 5573 (+) | 55.56(+) | **48.3667** |
| 30 | 11-Tumors | 6239 (+) | 6221 (+) | 6223 (+) | 6223 (+) | 6180 (+) | 6147 (+) | **141.0**(−) | 416.1667 |
| 31 | MLL | 6205 (+) | 6149 (+) | 6117 (+) | 6205 (+) | 6079 (+) | 6089 (+) | 60.3 (+) | **41.7** |
| 32 | Lung Cancer 2 | 6272 (+) | 6220 (+) | 6123 (+) | 6223 (+) | 6132 (+) | 6117 (+) | **74.52**(−) | 109.9 |
| 33 | CML treatment | 6246 (+) | 6143 (+) | 6092 (+) | 6199 (+) | 6073 (+) | 6105 (+) | 30.24(+) | **21.68** |
| 34 | glioblastoma | 6263 (+) | 6194 (+) | 6137 (+) | 6245 (+) | 6132 (+) | 6114 (+) | 55.64(+) | **35.36** |
| 35 | AML prognosis | 6287 (+) | 6186 (+) | 6141 (+) | 6220 (+) | 6109 (+) | 6109 (+) | 61.72(+) | **43.96** |
| 36 | prostate cancer | 6260 (+) | 6159 (+) | 6127 (+) | 6235 (+) | 6076 (+) | 6102 (+) | 24 (+) | **18.88** |
| 37 | Ovarian Cancer | 7494 (+) | 7469 (+) | 7452 (+) | 7435 (+) | 7332 (+) | 7324 (+) | 43.63(+) | **23.9** |
| 38 | SMK_CAN_187 | 9958 (+) | 9915 (+) | 9965 (+) | 9943 (+) | 9935 (+) | 9954 (+) | 81.03(+) | **68.5667** |
| 39 | GLI_85 | 11088 (+) | 10919 (+) | 10928 (+) | 11029 (+) | 10867 (+) | 10889 (+) | **115.3**(−) | 225.1 |
| 40 | Breast Cancer | 12177 (+) | 12232 (+) | 12092 (+) | 12232 (+) | 12042 (+) | 12023 (+) | 84 (+) | **59.6667** |
| | Wilcoxon Test(+\| ≈ \|−) | 40 \| 0 \| 0 | 40 \| 0 \| 0 | 40 \| 0 \| 0 | 40 \| 0 \| 0 | 40 \| 0 \| 0 | 40 \| 0 \| 0 | 21 \| 0 \| 19 | — |
| | Friedman Test (Mean Rank) | 7.83 | 6.11 | 5.36 | 6.29 | 3.78 | 3.64 | 1.53 | **1.48** |

## II. RESULTS AND DISCUSSIONS

### A. Running time

Runtime is one of the most important criteria in evaluating the efficiency of a FS algorithm. In FS using EC methods, the running time of FS algorithms depends on various factors such as the running time of the methods used in the FS algorithm, as well as the running time of the fitness function evaluation. For example, in some FS methods, filter-based algorithms may be used to identify important features and reduce the dimensions of the search space. Using these algorithms increases the running time of the FS algorithm. It is important to note that the running time of the fitness function evaluation is dependent on the number of features in the dataset. The running time during evaluating solutions with a large number of features is longer than the running time of solutions with a small number of features. Therefore, the computational cost is significantly reduced if a FS algorithm can achieve the highest classification accuracy with the least number of features. To make a fair comparison, the number of evaluations for all algorithms is 6000.

Table VI shows the algorithms' results in average runtime in 30 independent runs per second for each dataset. The results indicate the high speed of SFE compared to the other algorithms. This is because of the capability of SFE to explore the search space using the non-selection operator for unimportant features and removing them from the main feature set. For example, there are 24,481 features in the Breast Cancer dataset, all six algorithms that continue the search process with about 12,000 features, while the SFE algorithm continues the search process with less than 100 features in most search phases. The results indicate that the SFE-PSO algorithm's running time in all datasets is slightly longer than the SFE algorithm. Figure 1 shows the average running time of different algorithms for all datasets; SFE and SFE-PSO has the lowest computational cost in compared to other algorithms.

The running time of the SFE algorithm shows the high speed of this algorithm compared to other algorithms. The SFE algorithm consumes an average of 159 minutes' computational cost for one run on all 40 datasets. In contrast, BDE, BGA, BPSO, DimReM-BDE, DimReM-BGA, and DimReM-BPSO algorithms require approximately 420 minutes of computational cost. The computational time consumed for the SFE-PSO algorithm is 9 minutes longer than that of the SFE algorithm. Therefore, since in the SFE and SFE-PSO algorithms, by starting the search process with the non-selection operator, a large number of unimportant features change into the non-selection mode, the running time of these algorithms is significantly shorter than the other six algorithms. This superiority is very significant in the dataset with a large number of features which can be seen in Table VI.

TABLE VI
RUNING TIME CONSUMED BY THE EIGHT ALGORITHMS ON THE 40 DATASETS (UNIT: S)

| No. | Dataset | BDE [43] | BGA [5] | BPSO [44] | DimReM-BDE [5] | DimReM-BGA [5] | DimReM-BPSO [5] | SFE | SFE-PSO |
|---|---|---|---|---|---|---|---|---|---|
| 1 | ORL | 541.2461(+) | 583.4642(+) | 544.4080(+) | 533.0367(+) | 528.3768(+) | 539.2996(+) | **274.7719**(−) | 307.4716 |
| 2 | Yale | 500.1364(+) | 491.0909(+) | 474.1818(+) | 442.8182(+) | 424.2273(+) | 411.0000(+) | **241.0767**(−) | 246.2473 |
| 3 | Colon | 283.9458(+) | 246.6358(+) | 248.5017(+) | 263.4985(+) | 283.9458(+) | 302.9982(+) | **203.5401**(−) | 210.2160 |
| 4 | SRBCT | 259.0920(+) | 279.9288(+) | 268.5830(+) | 272.3974(+) | 274.6476(+) | 264.8922(+) | **206.2531**(−) | 215.5863 |
| 5 | warpAR10P | 315.0313(+) | 310.9073(+) | 302.5929(+) | 311.6762(+) | 293.6838(+) | 297.8050(+) | **204.4825**(−) | 212.6028 |
| 6 | Lung Cancer 1 | 553.6767(+) | 543.3894(+) | 624.7057(+) | 566.1042(+) | 535.3923(+) | 539.2301(+) | **262.2813**(−) | 281.0564 |
| 7 | Lymphoma | 377.8539(+) | 350.6657(+) | 349.7745(+) | 368.9600(+) | 346.6688(+) | 350.7426(+) | **239.0331**(−) | 262.2968 |
| 8 | GLIOMA | 351.5166(+) | 442.2780(+) | 277.6318(+) | 258.5156(+) | 288.3924(+) | 257.9036(+) | **235.8901**(−) | 249.7273 |
| 9 | Leukemia 1 | 354.0600(+) | 330.2223(+) | 332.2564(+) | 374.9271(+) | 324.3148(+) | 326.7226(+) | **250.8541**(−) | 261.6521 |
| 10 | DLBCL | 389.2433(+) | 380.3789(+) | 375.1956(+) | 384.2798(+) | 365.1615(+) | 364.0092(+) | **244.6797**(−) | 248.8541 |
| 11 | bladder cancer | 290.9187(+) | 289.7900(+) | 286.8089(+) | 283.8560(+) | 272.7860(+) | 287.1854(+) | **221.7138**(−) | 235.1802 |
| 12 | 9-Tumors | 320.5015(+) | 302.1543(+) | 304.5801(+) | 317.9044(+) | 298.9058(+) | 302.3159(+) | **250.8441**(−) | 265.2514 |
| 13 | TOX_171 | 603.0840(+) | 590.4988(+) | 590.3512(+) | 593.5654(+) | 581.6480(+) | 590.3741(+) | **277.8532**(−) | 295.6514 |
| 14 | Brain_Tumor1 | 394.2107(+) | 371.7299(+) | 427.3627(+) | 416.1046(+) | 369.0812(+) | 399.1046(+) | **251.6583**(−) | 267.2031 |
| 15 | Prostate GE | 424.1751(+) | 450.2496(+) | 456.9916(+) | 427.5325(+) | 439.9723(+) | 441.6824(+) | **274.4060**(−) | 290.2154 |
| 16 | leukemia | 377.8539(+) | 350.6657(+) | 349.7749(+) | 368.9600(+) | 346.6688(+) | 350.7428(+) | **251.7532**(−) | 267.5201 |
| 17 | ALLAML | 368.2364(+) | 344.0183(+) | 344.7949(+) | 358.6048(+) | 343.5811(+) | 343.6836(+) | **217.2698**(−) | 242.1219 |
| 18 | CNS | 380.4685(+) | 347.4508(+) | 375.7153(+) | 477.1344(+) | 348.2764(+) | 351.6070(+) | **223.4291**(−) | 246.1087 |
| 19 | ALL-AML-3 | 440.2320(+) | 422.6172(+) | 381.1674(+) | 489.5207(+) | 376.4393(+) | 431.2530(+) | **251.5603**(−) | 247.1452 |
| 20 | ALL-AML-4 | 469.0674(+) | 441.9734(+) | 442.2845(+) | 390.1674(+) | 376.7197(+) | 365.1251(+) | **245.5339**(−) | 266.5186 |
| 21 | Leukemia 2 | 426.1322(+) | 406.2893(+) | 431.5671(+) | 422.5816(+) | 395.4409(+) | 418.8220(+) | **251.0294**(−) | 270.3005 |
| 22 | childhood ALL | 467.9675(+) | 460.8471(+) | 451.0124(+) | 482.6325(+) | 464.3200(+) | 460.9524(+) | **271.4392**(−) | 283.3181 |
| 23 | Carcinom | 479.2894(+) | 471.0015(+) | 465.1265(+) | 479.2029(+) | 470.2564(+) | 472.2954(+) | **223.8957**(−) | 235.5214 |
| 24 | nci9 | 395.6600(+) | 499.8097(+) | 359.5930(+) | 392.7161(+) | 370.1463(+) | 414.9878(+) | **209.2154**(−) | 218.9003 |
| 25 | arcene | 1386.102(+) | 1340.037(+) | 1401.403(+) | 1392.215(+) | 1368.321(+) | 1375.573(+) | **255.1887**(−) | 269.2514 |
| 26 | orlraws10P | 561.4201(+) | 690.8217(+) | 521.7263(+) | 547.4365(+) | 527.7035(+) | 580.4538(+) | **261.0213**(−) | 275.8608 |
| 27 | Brain_Tumor2 | 402.2516(+) | 398.4953(+) | 399.8541(+) | 405.0210(+) | 401.1234(+) | 408.0094(+) | **245.8012**(−) | 258.6450 |
| 28 | Leukemia 3 | 540.9726(+) | 462.8705(+) | 534.8302(+) | 530.9527(+) | 469.2992(+) | 486.2307(+) | **241.1309**(−) | 255.8521 |
| 29 | CLL SUB_111 | 838.0785(+) | 803.5214(+) | 801.0213(+) | 784.8991(+) | 758.3350(+) | 789.2531(+) | **243.9980**(−) | 251.5293 |
| 30 | 11-Tumors | 1372.096(+) | 1223.859(+) | 1222.165(+) | 1572.976(+) | 1565.088(+) | 1362.643(+) | **252.1077**(−) | 273.8871 |
| 31 | MLL | 554.3502(+) | 480.3871(+) | 568.1775(+) | 526.9200(+) | 488.3199(+) | 515.9554(+) | **229.1854**(−) | 239.2937 |
| 32 | Lung Cancer 2 | 1593.254(+) | 1513.923(+) | 1532.202(+) | 1734.090(+) | 1562.6366(+) | 1723.71(+) | **265.1183**(−) | 299.6739 |
| 33 | CML treatment | 362.4804(+) | 346.4675(+) | 327.8606(+) | 352.1100(+) | 335.6387(+) | 352.6826(+) | **202.9308**(−) | 205.9978 |
| 34 | glioblastoma | 442.6616(+) | 418.4480(+) | 453.7888(+) | 431.3714(+) | 380.8300(+) | 458.6237(+) | **204.8273**(−) | 209.7009 |
| 35 | AML prognosis | 363.2683(+) | 332.4697(+) | 328.2568(+) | 346.6846(+) | 303.0070(+) | 332.1513(+) | **201.6439**(−) | 215.4942 |
| 36 | prostate cancer | 316.5398(+) | 303.8168(+) | 353.2002(+) | 364.8926(+) | 342.3289(+) | 325.8817(+) | **215.6388**(−) | 223.3617 |
| 37 | Ovarian Cancer | 3210.561(+) | 3111.514(+) | 3118.526(+) | 3110.967(+) | 3105.965(+) | 3109.001(+) | **276.3198**(−) | 291.6594 |
| 38 | SMK_CAN_187 | 2328.012(+) | 2319.251(+) | 2306.294(+) | 2311.351(+) | 2318.351(+) | 2343.012(+) | **236.9516**(−) | 238.6916 |
| 39 | GLI_85 | 917.1863(+) | 890.7940(+) | 895.1259(+) | 895.9701(+) | 949.3433(+) | 882.7051(+) | **224.4827**(−) | 228.5900 |
| 40 | Breast Cancer | 1410.124(+) | 1386.253(+) | 1330.001(+) | 1382.025(+) | 1389.201(+) | 1342.012(+) | **244.6513**(−) | 251.3931 |
| | Wilcoxon Test(+ | ≈ | −) | 40 | 0 | 0 | 40 | 0 | 0 | 40 | 0 | 0 | 40 | 0 | 0 | 40 | 0 | 0 | 40 | 0 | 0 | 0 | 0 | 40 | − |
| | Friedman Test (Mean Rank) | 6.99 | 5.23 | 5.28 | 6.15 | 4.21 | 5.15 | **1.02** | 1.98 |

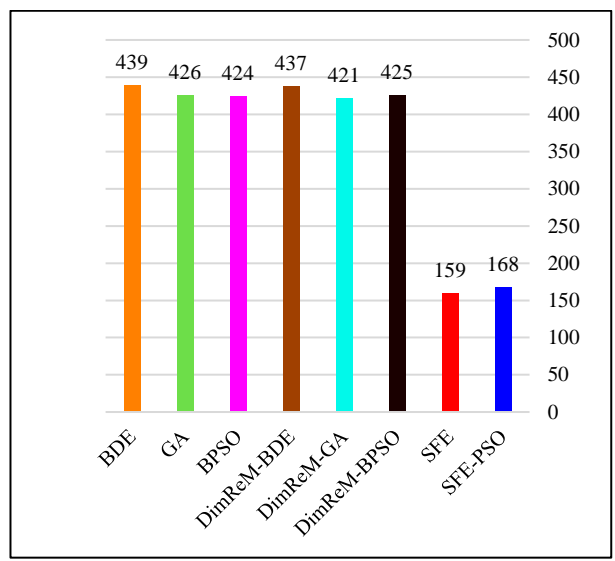

Fig. 1. The average of training time consumed by the eight algorithms on the 40 datasets (unit: m).

## III. FURTHER ANALYSIS OF THE RESULTS OF THE EXPERIMENTS

In this section, we analyze the results of the experiments performed on classification accuracy criteria, the number of selected features, and the behavior of the SFE and SFE-PSO algorithms in the search process.

### C. Classification accuracy

The significant reason for the superiority of SFE and SFE-PSO algorithms against the other algorithms based on the average accuracy is the use of the non-selection operator for changing the status of the irrelevant features to non-selection mode and the use of the selection operator to add the relevant features to the subset of optimal features. In fact, in high-dimensional datasets, especially in microarray data, many features are irrelevant to the classification task. Therefore, if we can identify them with mechanisms and put them in non-selection mode, we can achieve the expected efficiency. This mechanism is the non-selection operator in the SFE algorithm.

In addition, the algorithm escapes from the local minimums and achieves the best solutions due to the stochastic/random nature of the operators. For example, Figure 2 shows the search process of the SFE algorithm in one run on the Colon dataset.

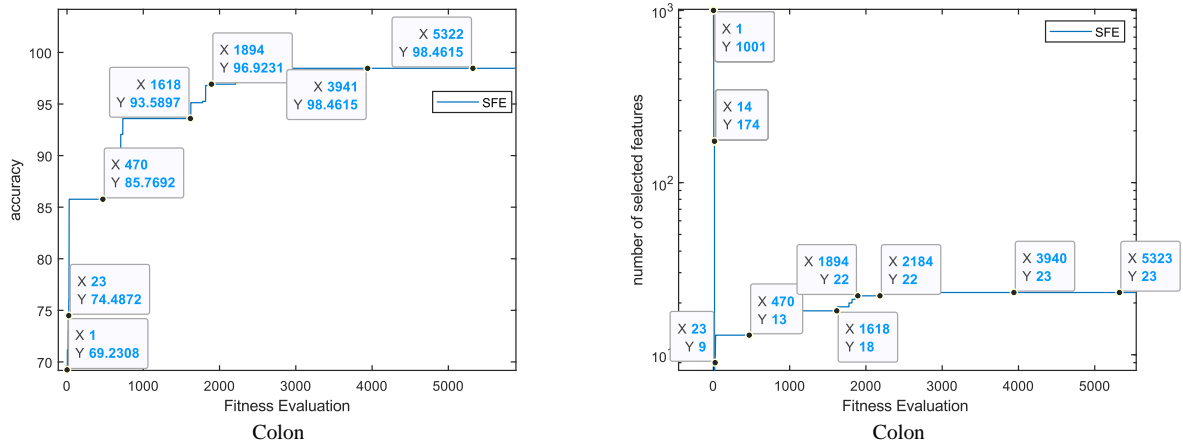

Fig. 2. The search process of the SFE algorithm in one run on the Colon dataset.

The figure on the right shows the number of features during run. The algorithm starts its search process with 1001 features; in the 23rd iteration, the number of selected features reaches nine features (using the no-selection operator). The same thing decreases the algorithm's computational cost in the continuation of the search process, and the important features (with the selection operator) are added to the features set. The figure on the left shows the convergence criterion of the algorithm in this execution. The algorithm has achieved a classification accuracy of 96.92 in 1894 iterations by starting the search from accuracy of 69.23 in iteration 1.

On the opposite point, EC methods do not have such mechanisms in the search process, so after starting the search process and after several iterations, they stop at one of the local optima of the problem, and the expected efficiency is not achieved. For example, Figure 3 shows the search process of the BPSO algorithm in one run on the Colon dataset.

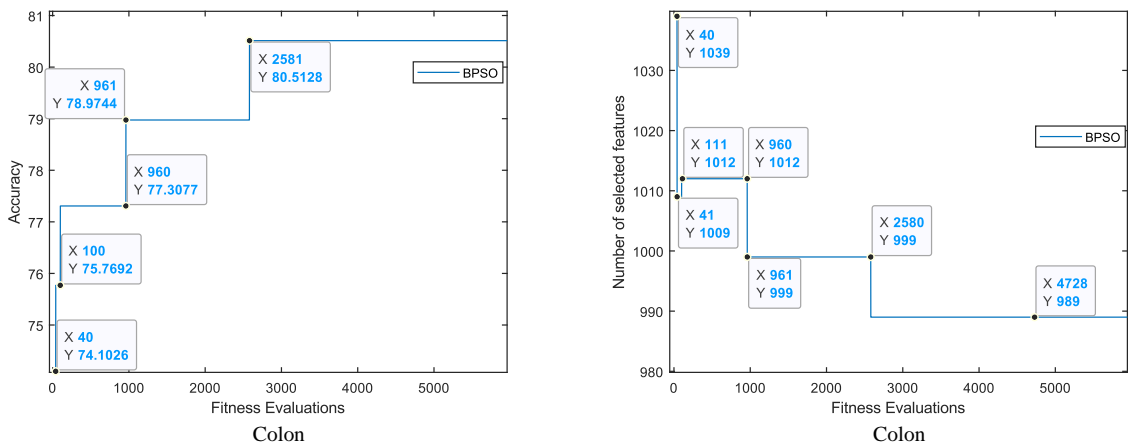

Fig. 3. The search process of the BPSO algorithm in one run on the Colon dataset.

The figure on the right displays the number of features during one run. In iteration 40, the BPSO algorithm selected 1039 features; in iteration 41, the number of selected features reached 1009. Next, the algorithm finishes the search process with 989 features. In fact, the algorithm has not been able to select important features. The figure on the left shows the convergence criterion of the algorithm in this execution.

The algorithm has achieved a classification accuracy of 74.10 in 40 iterations. In the continuance of the search process, several improvements have been made in the classification accuracy criterion. However, finally, it stopped in the 2581 iterations of the algorithm in a local optimum. The results of the SFE and BPSO algorithms on SMK_CAN_187, CNS, and CLL SUB_111 datasets can also be seen in Figure 4 and Figure 5.

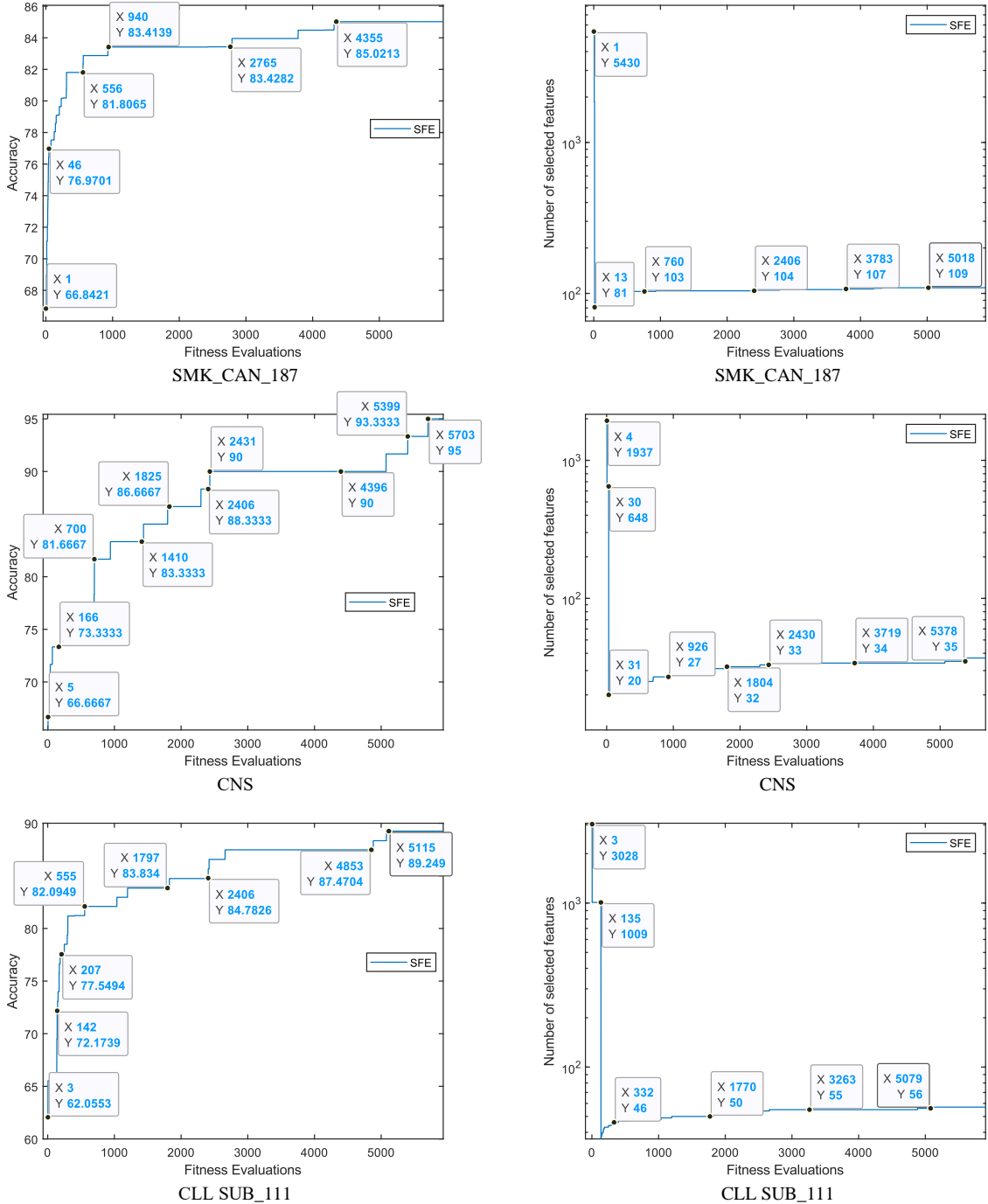

Fig. 4. The search process of the SFE algorithm in one run on the SMK_CAN_187, CNS, and CLL SUB_111 datasets.

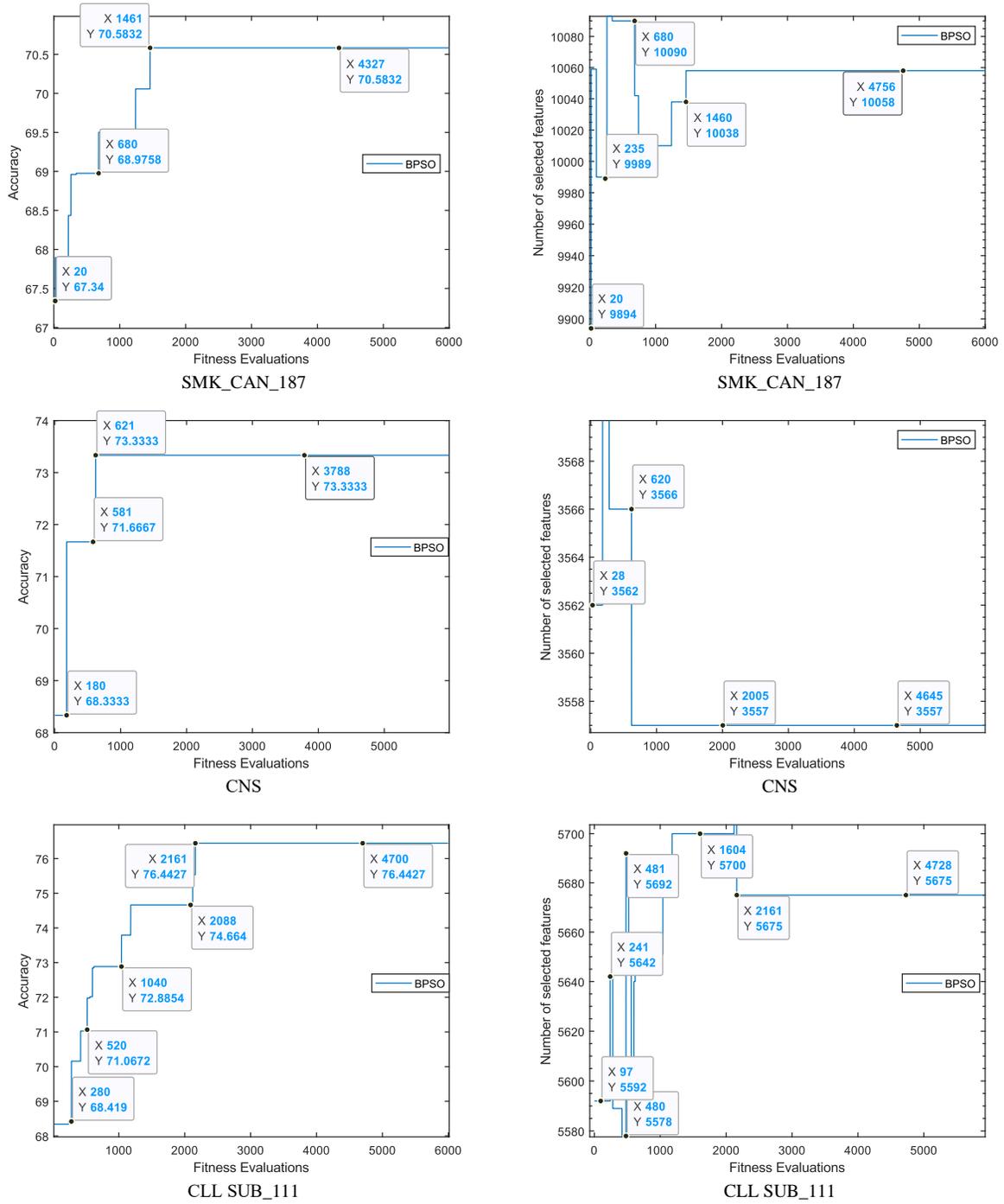

Fig. 5. The search process of the BPSO algorithm in one run on the SMK_CAN_187, CNS, and CLL SUB_111 datasets.

Moreover, in the SFE-PSO algorithm, the PSO continues the search process to find the best feature subset from the feature set determined by the SFE algorithm. On the contrary, the three algorithms, BDE, BGA, and BPSO, stop at one of the local optimal points of the high-dimensional data, and because of their search strategy, they cannot reach better classification accuracy during the search process. In addition, three algorithms, DimReM-DBE, DimeRem-BGA, and DimReM-BPSO, are not able to perform classification with high accuracy because of their slow process of removing irrelevant features. Other state-of-the-art EC-based feature selection algorithms, i.e., HFS-C-P, SaWDE, and SM-MOEA, had competitive results with the proposed algorithms. However, the results of the Wilcoxon and Friedman tests showed the superiority of the SFE-PSO algorithm over these algorithms.

### D. Number of selected features

The reason for the significant superiority of SFE and SFE-PSO algorithms compared to BDE, BGA, and BPSO, DimReM-BDE, DimReM-BGA, and DimReM-BPSO algorithms in the criterion of the number of selected features is the use of the non-selection operator. In fact, the SFE algorithm changes the status of unimportant features from selection mode to non-selection mode using the non-selection operator. Therefore, the algorithm finds solutions that have the least number of features. Also, relevant and important features are added to the optimal subset of features at different stages of the search process using the selection operator. On the opposite point, the three algorithms, BDE, BGA, and BPSO, have almost half of the features of the data in their solutions in their entire search process, most of which are unimportant features. Therefore, since there are no mechanisms to identify unrelated features in these methods, they practically cannot find a smaller number of features. Also, in the three algorithms DimReM-BDE, DimReM-BGA, and DimReM-BPSO, the process of identifying irrelevant features and removing them is done slowly, and finally, these algorithms cannot obtain solutions with fewer features.

### E. The behavior of the SFE and SFE-PSO algorithms in the search process

Viewing Figure 6, which is the convergence diagram of algorithms in different datasets, gives us important information about the behavior of algorithms in the search process. Observing the behavior of SFE and SFE-PSO algorithms in some datasets such as ORL, Yala, Carcinom, TOX_171, and Carcinom shows that in these datasets, the SFE algorithm has Poor performance for reasons such as the inappropriate value of the $UN$ and $SN$ parameters. for this reason, the algorithms were not able to achieve better accuracies in the early stages of the search process. Meanwhile, the SFE algorithm has been able to obtain better solutions in some datasets, such as TOX_171, in the continuation of the search process. On the other hand, in some datasets, such as Yale, the algorithm has not significantly improved classification accuracy even until the last iteration. Observing the behavior of the SFE-PSO algorithm shows that after 2000 iterations and continuing the search process with PSO algorithm, this algorithm has achieved a significant classification accuracy improvement. The reason for this is the ability of the PSO algorithm to select features in low-dimensional data. In fact, after the PSO algorithm continues the search process in the search space reduced by the SFE algorithm, it obtains better solutions. This issue can be seen from the algorithm's results in datasets such as Yale, lymphoma, and TOX_171.